\documentclass{article}

\usepackage{microtype}
\usepackage{graphicx}
\usepackage{subcaption}
\usepackage{booktabs} %
\usepackage{xspace}
\usepackage[dvipsnames]{xcolor}

\usepackage{hyperref}

\usepackage[accepted]{sysml2019}

\usepackage{letltxmacro}
\LetLtxMacro\oldttfamily\ttfamily
\DeclareRobustCommand{\ttfamily}{\oldttfamily\csname ttsize\endcsname}
\newcommand{\setttsize}[1]{\def\ttsize{#1}}%

\setttsize{\footnotesize}

\usepackage{fancyvrb}

\def\Snospace~{\S{}}
\def\Fnospace~{\mbox{Fig.\hspace{0.25em}}}
\def\Tnospace~{\mbox{Tab.\hspace{0.25em}}}
\def\Enospace~{\mbox{Equation\hspace{0.25em}}}

\newcommand{\eat}[1]{}

\usepackage{cleveref}

\newcommand{\ghold}{\textsc{GH17}\xspace}
\newcommand{\ghnew}{\textsc{GH19}\xspace}
\newcommand{\tlc}{\textsc{AnonSys}\xspace}
\newcommand{\pypi}{\textsc{PyPI}\xspace}
\newcommand{\gh}{\textsc{GitHub}\xspace}
\newcommand{\companyname}{\textsc{CompanyX}\xspace}

\newcommand{\mynote}[1]{\par\noindent\colorbox{lightgray}{\parbox{\linewidth}{#1}}}
\newcommand{\fp}[1]{\textcolor{red}{\mynote{FP:~#1}}}
\newcommand{\mi}[1]{\textcolor{red}{\mynote{MI:~#1}}}
\newcommand{\yz}[1]{\textcolor{red}{\mynote{YZ:~#1}}}

\newcommand{\kk}[1]{\textcolor{red}{\mynote{KK:~#1}}}

\newcommand{\ifp}[1]{\colorbox{lightgray}{\textcolor{MidnightBlue}{FP:~#1}}}

\newcommand{\mysim}{\raise.17ex\hbox{$\scriptstyle\sim$}}

\newcommand{\sklearn}{\textsc{Scikit-Learn}\xspace}
\newcommand{\numpy}{\textsc{Numpy}\xspace}
\newcommand{\matplotlib}{\textsc{Matplotlib}\xspace}
\newcommand{\pandas}{\textsc{Pandas}\xspace}
\newcommand{\scipy}{\textsc{Scipy}\xspace}
\newcommand{\seaborn}{\textsc{Seaborn}\xspace}
\newcommand{\theano}{\textsc{Theano}\xspace}
\newcommand{\nolearn}{\textsc{NoLearn}\xspace}
\newcommand{\keras}{\textsc{Keras}\xspace}
\newcommand{\nimbusml}{\textsc{NimbusML}\xspace}
\newcommand{\tensorflow}{\textsc{Tensorflow}\xspace}
\newcommand{\mxnet}{\textsc{Mxnet}\xspace}
\newcommand{\os}{\textsc{Os}\xspace}
\newcommand{\sys}{\textsc{Sys}\xspace}
\newcommand{\cv}{\textsc{OpenCV}\xspace}
\newcommand{\PIL}{\textsc{Pillow}\xspace}
\newcommand{\torch}{\textsc{PyTorch}\xspace}
\newcommand{\caffe}{\textsc{Caffe}\xspace}
\newcommand{\nltk}{\textsc{NLTK}\xspace}
\newcommand{\re}{\textsc{Regular Expression}\xspace}
\newcommand{\pylab}{\textsc{Pylab}\xspace}
\newcommand{\statsmodels}{\textsc{StatsModels}\xspace}
\newcommand{\requests}{\textsc{Requests}\xspace}
\newcommand{\xgboost}{\textsc{XGBoost}\xspace}
\newcommand{\gensim}{\textsc{Gensim}\xspace}
\newcommand{\tqdm}{\textsc{Tqdm}\xspace}
\newcommand{\sqlalchemy}{\textsc{SQLAlchemy}\xspace}
\newcommand{\bs}{\textsc{bs4}\xspace}
\newcommand{\lasagne}{\textsc{Lasagne}\xspace}

\newcommand{\code}[1]{\textcolor{MidnightBlue}{#1}}

\usepackage{microtype}
\usepackage{listings}
\usepackage{url}
\usepackage{balance}
\usepackage{authblk}
\usepackage[accepted]{sysml2019}

\sysmltitlerunning{Data Science Through the Looking Glass}
\graphicspath{{figures/}} 

\renewcommand{\mynote}[1]{\eat{#1}}

\newcommand{\AF}[1]{\textcolor{magenta}{\mynote{AF:~#1}}}
\newcommand{\SK}[1]{\textcolor{orange}{\mynote{SK:~#1}}}
\newcommand{\wag}[1]{{\textbf{{{WAG:~\emph{#1}}}}}}

\usepackage{enumitem}
\setlist{leftmargin=*,itemsep=-.2em,topsep=0em}

\renewenvironment{thebibliography}[1]{
  \begin{oldthebibliography}{#1}
    \setlength{\itemsep}{0em}
    \setlength{\parskip}{0em}
}
{
  \end{oldthebibliography}
}

\begin{document}

\twocolumn[
    
\sysmltitle{Data Science Through the Looking Glass \\ and What We Found There}

\sysmlsetsymbol{equal}{*}

\begin{sysmlauthorlist}
  \sysmlauthor{Fotis Psallidas}{m}
  \sysmlauthor{Yiwen Zhu}{m}
  \sysmlauthor{Bojan Karlas}{m,eth}
  \sysmlauthor{Matteo Interlandi}{m}
  \sysmlauthor{Avrilia Floratou}{m}
  \sysmlauthor{Konstantinos Karanasos}{m}
  \sysmlauthor{Wentao Wu}{m}
  \sysmlauthor{Ce Zhang}{eth}
  \sysmlauthor{Subru Krishnan}{m}
  \sysmlauthor{Carlo Curino}{m}
  \sysmlauthor{Markus Weimer}{m}
  
  \end{sysmlauthorlist}
  
  \sysmlaffiliation{m}{Microsoft, Redmond, Washington, USA}
  \sysmlaffiliation{eth}{ETH, Zurich, Zurich, Switzerland}
  \sysmlcorrespondingauthor{Fotis Psallidas}{f.p@ms.com}
\vskip 0.3in

\begin{abstract}
The recent success of machine learning (ML) has led to an explosive growth both in terms of new systems and algorithms built in industry and academia, and new applications built by an ever-growing community of data science (DS) practitioners. This quickly shifting panorama of technologies and applications is challenging for builders and practitioners alike to follow. 
In this paper, we set out to capture this panorama through a wide-angle lens,
by performing the largest analysis of DS projects to date, focusing on questions that can help determine investments on either side.  Specifically, we download and analyze: (a)~over 6M Python notebooks publicly available on \gh, (b)~over 2M enterprise DS pipelines developed within \companyname, and (c) the source code and metadata of over 900 releases from 12 important DS libraries. The analysis we perform ranges from coarse-grained statistical characterizations to analysis of library imports, pipelines, and comparative studies across datasets and time. 
We report a large number of measurements for our readers to interpret, and dare to draw a few (actionable, yet subjective) conclusions on (a)~what systems builders should focus on to better serve practitioners, and (b)~what technologies should practitioners bet on given current trends. We plan to automate this analysis and release associated tools and results periodically.
\end{abstract}

\eat{
\SK{Reopening the title debate - "I feel we can more explicitly state analysis of 5M GitHub notebooks", for SEO as this work should be heavily cited.}

\AF{I think we need to emhasize more on the contrast between system builder and data scientist and thus motivate more strongly the goal of the study. Maybe we should say that system builders often lack complete understanding of data science workloads due to organizaotional boundaries or other reasons but they are still making crucial decisions that affect data scientists' work.}
\mi{I think for \tlc we have 29M pipelines, 2M are the unique ones (for \gh notebooks are 5M and unique 3M.)}

}
]

\printAffiliationsAndNotice{}

\section{Introduction}
\label{s:intro}

The ascent of machine learning (ML) to mainstream technology is in full swing: from academic curiosity in the 80s and 90s to core technology enabling large-scale Web applications in the 90s and 2000s to ubiquitous technology today. Given many conversations with enterprises, we expect that in the next decade most applications will be  {\em ``ML-infused''}, as also discussed in~\cite{cidrvision}.  This massive commercial success and academic interest are powering an unprecedented amount of engineering and research efforts---in the last two years alone we have seen over 23K papers in a leading public archive,\footnote{Per arXiV search: \url{https://bit.ly/2lJBoQu}.} and millions of publicly shared data science (DS) notebooks corresponding to tens of billions of dollars of development time.\footnote{Per COCOMO software costing model \cite{cocomo} applied to the \gh datasets we discuss in this paper.}

As our team began investing heavily both in building systems to support DS and leveraging DS to build applications, we started to realize that the speed of evolution of this field left our system builders uncertain on what DS practitioners needed (e.g., {\em Are practitioners shifting to using only DNNs?}).  On the other end, as DS practitioners we were equally puzzled on which technologies to learn and build upon (e.g., {\em Shall we use \tensorflow or \torch?}). As we interviewed experts in the field, we got rather inconsistent opinions. 

We thus embarked in a (costly) fact finding mission, consisting of large data collection and analysis, to construct a better vantage point on this shifting panorama.  As more and more of the results rolled in, we realized that this knowledge could serve the community at large, and compiled a short summary of the key results in this paper. 

The goal of this paper is to create a data analysis-driven bridge between system builders and the vast audience of data scientists and analysts that will be using these systems to build the next one million {\em ML-infused} applications.\footnote{Per estimates reported in~\cite{cidrvision}.} 
To this end, we present the largest analysis of DS projects to date: (a)~6M Python notebooks, publicly shared on \gh---representative of OSS, educational, and self-learning activities;  (b)~2M DS pipelines professionally authored in \tlc within \companyname, a planet-scale Web company---representative of the state of DS in a very mature AI/ML ecosystem; and (c)~an analysis of over 900 releases of 12 important DS libraries---this captures growth, popularity, and maturity of commonly used tools among practitioners.
  
The diversity and sheer size of these datasets enable multiple dimensions of analysis. In this paper, we focus on extracting insights from dimensions that are most urgent for the development of systems for ML, and for practitioners to interpret adoption and support trends:

\begin{itemize}[itemsep=-0.5em]
\item {\em Landscape} (\autoref{s:coarsestats}) provides a bird's-eye view on the volume, shape, and authors of DS code---hence, provides the aggregate landscape of DS that systems aim to support.
    
\item {\em Import analysis} (\autoref{s:imports}) provides a finer-grained view of this landscape, by analyzing the usage of DS libraries both in isolation and in correlation. As such, it (a)~sheds light on the functionality that data scientists rely on and systems for ML need to focus on, and (b)~provides a way to prioritize efforts based on the relative usage of libraries.

\item {\em Pipeline analysis} (\autoref{s:pipelines}) provides an even finer-grained view by analyzing operators (e.g., learners, transformers) and the shape (e.g., \#operators) of well-structured and, often, manageable and optimizable DS code~\cite{schelter2017automatically,cidrvision}.

\item {\em Comparative analysis} (\autoref{s:modules} and throughout the paper) is performed, when possible, across time and datasets. In particular, we compare: (a)~the evolution of 12 widely used Python libraries for DS across all their releases, investigating their code and API stability; (b)~statistics for Python notebooks from \gh between 2017 and 2019; and (c)~\sklearn pipelines from \gh with \tlc DS pipelines, studying similarities/differences of public notebooks and mature enterprise DS pipelines.
\end{itemize}

In addition to reporting objective measures, we feel compelled to provide more subjective and speculative interpretations of the results---we call these Wild Actionable Guesses (WAGs). In order to clearly distinguish between these two classes, we will systematically mark our speculations with the following typography: \wag{A speculative example}. 

As an example, comparing \gh in 2017 and in 2019 we observe a $3\times$ increase in the number of DS libraries used. Interestingly, however, the most popular libraries gained even more prominence ($+3\%$ for the top-5). Moreover, the top-7 operators in \tlc cover 75\% of the 2M pipelines we collected from \companyname. The wild actionable guess we derive from this is: \wag{System builders can focus their optimization efforts on a small number of core operations, but need mechanisms to support an increasingly diverse tail of libraries/functionalities.}

Collectively these datasets provide us with a unique 
vantage point on this important field.  We share the key 
findings in this paper in the form of aggregated statistics 
and insights we derived. However, to fully empower 
practitioners along systems and ML audiences at large, we 
are in the  process of releasing the public datasets (raw and 
processed versions) and our analysis tools. This will enable validation 
of our process, level the playing field, and enable many more
interesting questions to be asked and answered. We plan to 
maintain a continuous data collection, so that this analysis 
could become a ``living'' one, with continuously updated 
statistics.

Our goal with this work is to make a small contribution to help 
our community to converge to a common understanding of the 
``shape'' of this field, so that {\em we can all 
descend a shared gradient towards a global optimum}.

\eat{

\AF{Again maybe we should discuss more clearly  the gap between system builders and pratcitioners. Giving a couple of conrete examples upfront where our analysis can help system builders would be helpful. These examples should mention the type of system and constraints clearly so that the reviewers feel they are realistic. The current questions in the 2nd paragraph are towards that direction but I think we need to better ground them and also emphasize the mismatch.}
\AF{The intro  does not contain any major findings/insights from the paper but mostly a description of the various components of the study. Should we add three high-level major points towards the end as major findings/conclusions from the work?}

}

\section{Corpora, ETL, and Limitations}
\label{s:corpora}

For our analysis we leverage all publicly available notebooks from \gh, a large dataset of data science pipelines  from within \companyname, and OSS packages available from \pypi~\cite{pypi}. In this section, we describe each data source, and the ETL processes used to prepare data for analysis. Here, we report on overall data volumes for context. Detailed analysis is deferred to later sections.

\textbf{\gh.} In our analysis we use two corpora of publicly available notebooks on \gh, referred to as \ghold and \ghnew. Both consist of notebooks available at the \texttt{HEAD} of the \texttt{master} branch of all public repositories at the time of download: 1.2M notebooks (270G compressed) in July 2017 for \ghold~\cite{rule:2018,rule:2018:collection} and 5.1M notebooks (1.3T compressed) in July 2019 for \ghnew. 

\textbf{\tlc Telemetry.} Our second dataset is representative of the output of a mature data science community within \companyname ---a large scale web company. The underlying system, \tlc, has been developed and used in production for over a decade. We obtained access to a telemetry database from 2015 to 2019, containing over 88M reporting events. While many users opted-out of reporting telemetry, this large scale sample is representative of DS activities within \companyname, and provides an alternative (to \gh) vantage point.

\textbf{PyPI.} Our analysis in~\autoref{s:coarsestats} will reveal that both \ghnew and \ghold are dominated by notebooks authored in Python. We thus reuse our parsing/analysis infrastructure to inspect not just DS notebooks but also core libraries they use. We select the most popular libraries in \ghnew and few emerging ones to download all metadata and source code for all their releases from \pypi. This corresponds to 12 libraries, totalling 936 releases, 490K files, 2.3M functions, and 266K classes.

\eat{
\kk{Maybe call this technical details/technical challenges?}
\kk{Knowing the effort it took to do this analysis, these paras really understate the effort. Maybe mention data sizes, DB size/number of instances, challenges that we overcame, etc. It is very important to show here how challenging it was to pull this off.}
}

\textbf{ETL.} We built an in-house system to support code analytics at the scale described above. While the presentation of this system is beyond the scope of this paper, here we briefly describe the extract-transform-load (ETL) process that transforms the raw data provided from each source to a form that we can base our analysis. 
We employee a parallel crawler, based on \cite{rule:2018,rule:2018:collection}.  Upon downloading, we parse the \code{nbformat} format of each notebook, discard malformed notebooks, and upload the valid ones to a PostgreSQL database. We include all metadata (kernel, language, and nbformat versions), and cell-level information (type, order, source code). \tlc pipelines are similarly processed, and \pypi libraries are also added to the DB including all metadata and source code.  
We then perform several extraction passes, where the Python code is parsed and analyzed extracting different features (e.g., which libraries are imported in each code cell)---the features are also stored in the DB.  As a result, each statistic in this paper can be achieved by a (more or less complex) SQL query. We are evaluating the feasibility to release the raw DB, the ETL code, and those queries.

\textbf{Limitations.} A comprehensive analysis of all DS activities in the industry is an Herculean task beyond the ambitions of this paper. Here, we take a first stab to construct a representative dataset and some initial measurements. Data availability and parsing/analysis facilities at our disposal pushed us to focus on Python as a programming language, interactive notebooks (instead of all Python programs), and specifically publicly released notebooks. This certainly introduced several biases (e.g., we are more likely to include ``learning code'' than commercial DS code). To compensate we included the \tlc dataset. However here as well we carry deep biases: our \tlc dataset is limited by users opt-out choices, and this only captures the DS patterns typical for the use of \tlc at \companyname, which might differ from other larger companies, and certainly not representative of less mature/smaller players. Finally the sheer volume of data forces us to limit the manual inspections for the interpretation of our results.  

\eat{
\kk{See if GH17/GH19 looks better than GITHUB2017/2019 as it is shorter, up to you.}
\fp{done}
}

\eat{
\AF{I think this section should include some numbers to highlight the dataset sizes (e.g., number of notebooks in github, etc)-- I'm aware that section 4 gives all these details but I think we need to make a case earlier in the paper that we really looked into a lot of data and this seems to be a section where the reviewer would expect some info on scale}
\fp{done}
}

\eat{
\AF{Should we mention python vesrions here? Earlier to latest?}
\textbf{Python Distributions.} Along PyPI, we also downloaded all Python releases (excluding candidate or beta releases). From this source, we extracted the APIs of built-in modules that we use in our Module and API analysis (\autoref{s:modules,s:apis}, respectively). 

\fp{we do not analyze python distributions in this analysis. so I removed this part.}
}

\eat{
\SK{We should qualify "most frequently used".}
\fp{done.}
}

\eat{
\SK{We should set some context on why we are focusing on Python.}
\fp{done, when introducing pypi}
}

\eat{
\kk{Is 2019 only 2019 or does it include previous years too?}
\fp{handled}
}

\eat{
\kk{It is not clear to me how PyPI was used. Did we use something like the most popular packages from PyPI or the metadata of all of them? What did the metadata include? Then "associated source code (all available releases) from PyPI of the most frequently used packages in Python notebooks"---what was the source code used for; are the Python notebooks here the ones from github or something else?}
\fp{dealt in new text}
}

\eat{
\AF{Does it make sense to mention where each dataset is used and connect to follow-up sections in the paper? For example GIthub is used for import analysisn in section X, etc? Are all datasets used for all types of analysis?}
\fp{addressed}
}
\section{Landscape}
\label{s:coarsestats}

\eat{
\kk{Since we know the number of notebooks, below I'd just give percentages---clearer and that's what we care about after all.}
\fp{I will give this a shot if we have time.}
}

\eat{
backing up exact numbers before compressing
1,241,323 notebooks in \ghold
5,121,579 notebooks in \ghnew

34,696,693 cells in \ghold
148,039,575 cells in \ghnew

22,371,365 code cells in \ghold
98,424,526 code cells in \ghnew

1,485,364 empty code cells in \ghold
6,369,993 empty code cells in \ghnew

21,040 empty notebooks in \ghold
89,059 empty notebooks in \ghnew
}

We start our analysis by providing a birds' eye view of the landscape of data science through coarse-grained statistics from \gh notebooks. \autoref{tab:stats} presents an analysis of \gh notebooks to reveal the volume of (a) notebooks, cells, and code cells; (b) languages associated with them; and (c) their users. We present similar analysis for \tlc pipelines in~\autoref{s:pipelines} and for \pypi in~\autoref{s:modules}.

\begin{table}
\scalebox{0.8}{
\begin{tabular}{|ll|r|r|r|}
\hline 
{\bf Dataset} & & {\bf \ghold} & {\bf \ghnew} & {\bf Change} \\ 
\hline 
\hline
Notebooks & Total & 1.2M & 5.1M & $4.1\times$ \\ 
\hline 
 & Empty & 21K (1.6\%) & 89K (1.7\%) & $4.2\times$ \\ 
\hline 
 & Deduped & 816K (66\%) & 3M (59\%) & $3.7\times$ \\ 
\hline 
Cells & Total & 34.6M  & 148M & $4.3\times$ \\ 
\hline 
 & Empty & 1.5M (4.3\%) & 6.4M (4.32\%)& $4.3\times$ \\ 
\hline 
 & Code & 22M (64\%) & 98.4M (67\%) & $4.4\times$ \\ 
\hline 
 & Deduped & 9M (26\%) & 36M (24.3\%) & $4\times$ \\ 
\hline 
Language & Python & 1M (84\%) & 4.7M (92\%) & $4.7\times$ \\ 
\hline 
 & Other & 0.2M (16\%) & 0.4M (8\%) & $2\times$ \\ 
\hline 
User  & Unique & 100K & 400K & $4\times$ \\ 
\hline 
\end{tabular}  \\ 
}
\caption{Overall statistics for \gh.}
\label{tab:stats}
\vspace{-4mm}
\end{table}

We clean the raw data in several steps. First, we remove malformed notebooks. This leaves 1.24M notebooks for \ghold (5.1M for \ghnew) in our database. We then remove 4.3\% empty code cells (those containing only tabs, whitespace, and newlines) and 1.6\% empty notebooks (i.e., those made up of only empty cells). Next, we eliminate duplicate notebooks introduced by git forking, and notebook checkpointing\footnote{Users often uploads both a base notebooks, and the intermediate checkpoint files generated during editing.}. We retain 9M unique code cells for \ghold (36M for \ghnew). In subsequent sections, we will base our analysis on the full set of notebooks and code cells, and we will report when empty or duplicate code cells and notebooks have an effect on the results.

Overall, we see a roughly $4\times$ growth in most metrics between \ghold and \ghnew. Code cells growth outpaced notebooks growth (more cells per notebooks in \ghnew), while duplication was also increased in \ghnew. 

\textbf{Languages.} Looking at languages, we confirm one of our main hypothesis: Python grew its lead from 84\% in \ghold to 92\% in \ghnew. All other 100~languages combined grew by roughly $2\times$, well below Python's $4.7\times$. 
\wag{Python is emerging as a de-facto standard for DS; systems builders and practitioners alike should focus on Python primarily.}

\begin{figure}[t!]
	\vspace{-2mm}
	\centering
\includegraphics[width=0.9\columnwidth]{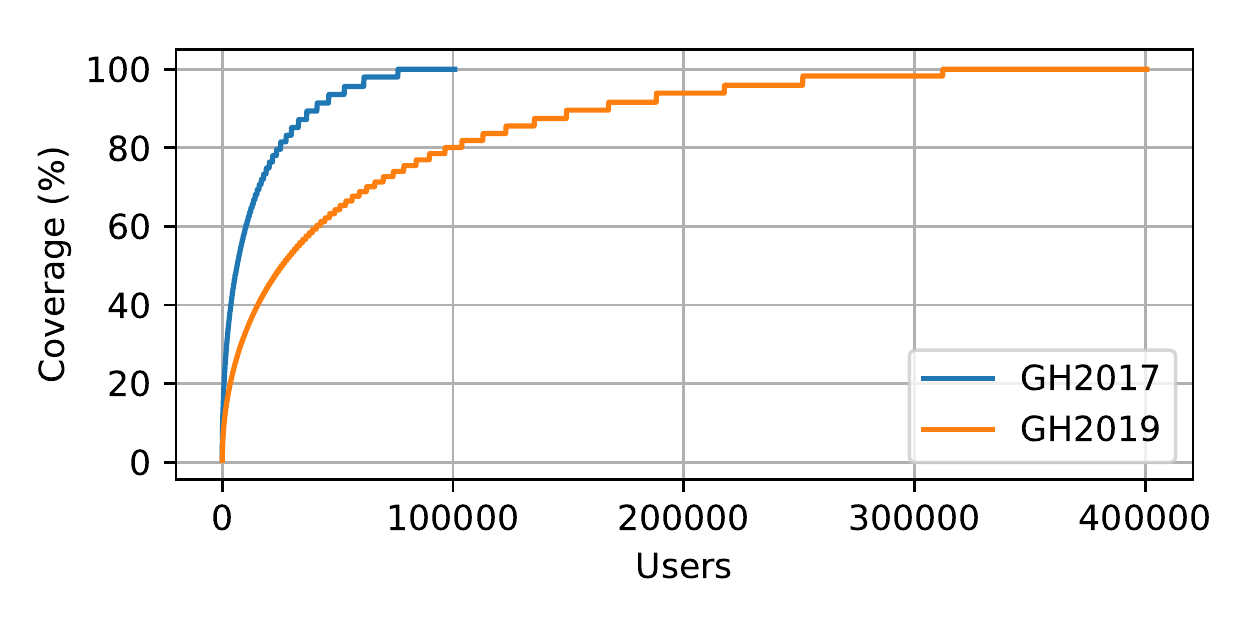}
	\vspace{-5mm}
	\caption{Coverage on \#notebooks with varying number of users. Users (x-axis) are ordered in descending order of \#authored notebooks. The absence of spikes on the left hand side indicates that no individual user dominates the collection of notebooks.}
	\label{f:ucov}
	\vspace{-5mm}
\end{figure}

\textbf{Users.} User growth is similarly paced at $4\times$, reaching an impressive 400K unique users in \ghnew. Note that on \gh, large organizations may appear as a single ``user''. Hence, this is a lower bound to the number of contributing individuals. The top 100 most prolific users, in this definition, have over 432 notebooks in \ghold (over 812 in \ghnew) each, yet no individual user provides more than 0.5\% of the notebooks in the corpus. This is best seen in~\autoref{f:ucov}. It shows the coverage of users on notebooks. Note the absence of spikes on the left hand side of the figure. Interestingly, the average number of notebooks per user has remained roughly the same between \ghold and \ghnew (i.e., 12.3 and 12.8 notebooks per user, respectively), with the standard deviation slightly increasing (i.e., 54.8 and 59.4, respectively).

\textbf{Code shape.} To get a better understanding of the shape of code in Python notebooks, we parsed (using \texttt{parso}~\cite{parso}) and analyzed every code cell to extract salient structural features. The 3M unique notebooks in \ghnew contain a combined 7.68B AST nodes, grouped into 13M functions, 1M classes, 10.7M \texttt{if} code blocks, 15.4M \texttt{for} loops, 0.78M \texttt{while} loops. Nesting is also present but reasonably rare with 3.3M functions in classes, 0.26M functions in functions, 3.1K classes in classes, and 3.8K classes in functions.
 Beside overall statistics we are curious to see how many notebooks or cells are \emph{linear}, i.e., contain no conditional statements, or \emph{completely linear}, i.e., contains neither conditionals statements nor classes or functions. At the notebook level for \ghnew, 28.53\% are completely linear while 35.11\% are linear. At the cell level, however, 83.04\% are completely linear while 87.52\% are linear. \ghold has similar characteristics for all metrics reported here.

Overall this analysis validates an interesting line of thinking: \wag{DS code is a mostly linear orchestration of libraries for data manipulation (+ some UDFs). It is thus amenable to be statically analyzed and transformed into declarative dataflows and optimized as such.}

\eat{
\AF{The title of the section is not very informative. Also a description of what is the goal of the section is missing. Shoudl we add in each subsection a 1-sentence describing why the analysis performed is useful? It will help better motivate the work.}
\fp{changed title and addressed the concern on purpose in introductory paragraph}
}

\eat{
\kk{Would be good to add here a high-level sentence to remind readers the purpose of this section.}
\fp{added in introductory paragraph}
}

\eat{
\kk{Why do we analyze only the GH stuff here and not the TLC ones too?}
\fp{At that time there was no description of TLC. I have dealt with TLC in the introductory paragraph and in subsequent paragraphs}
}

\eat{
\AF{How do we define distinct? If a cell is exactly the same as another except for a variable name change are these consider distinct?}
\fp{distinct: all have unique source code. for the second: yes, such code cells are different. aren't they?}
}

\eat{
\kk{Maybe call this "code complexity"?}
\fp{code complexity is well-established metric in loc counters. did not change the paragraph title to avoid the potential confusion.}
}

\eat{
\kk{Is it allowed to have multiple languages in a notebook? I think not, given there needs to be a unique kernel, but would be good to state.}
\fp{This opens up a very weird discussion that I do not want to get into. There is nothing stopping someone for creating notebooks with multiple languages. It's up to Jupyter or Azure data studio (and their equivalents) to support this. I am not sure if there is a product that offers this functionality and how they specify this in notebooks to extract and provide numbers about it.}
}

\eat{
\kk{Wondering if in millions would make the numbers more readable, as in 1.24M.}
\fp{changed}
}

\eat{
\kk{I would remove the "analysis of" from paragraph titles below.}
\fp{done}
}

\eat{
\kk{It is a bit tiring (and feels somehow handwavy) that we put a $\mysim$ next to all numbers. I think it is fine to drop when we give an okay precision (like 6.4\%).}
\fp{done}
}

\eat{
\AF{Why post-loading in the db? Before loading the number was different?}
\fp{added text in sec. 2 for this. in short, yes, if non-compliant with the nbformat specification they will be not considered further. these typically are wrong json files or files that were not downloaded because they were taken down between getting their metadata and downloading their acrual content (i.e., they result in html files with 404).}
}

\eat{
\AF{Would a table be more readable? Maybe have 3 columns year, \#notebooks, \# cells. SImilarly, we can add \#distinct cells, etc. Also agree with Kostas in presenting the number in millions.It's a bit hard to parse now.}
\fp{tried it. but I didn't like + it adds more space instead of removing even after trying to change the text. so I reverted the text back to before tables.}
}

\eat{
\kk{Para title says empty notebooks and cells, but give numbers only for the latter.}
\fp{fixed}
}

\eat{
816,239 distinct notebooks
3,009,880 distinct notebooks

9,147,754 distinct code cells
36,700,259 distinct code cells
}

\eat{
\kk{I'd flip the order below: empty notebooks are not that many and it is a boring reason (also state clearly that white notebooks/cells are considered non-distinct.)}
\fp{flipped. also addressed markus comment on quanitfying reasons for dedup}

}

\eat{
\begin{tabular}{|l|c|c|c|}
\hline
year &notebooks & cells & code cells \\ \hline
2017 & 1.24M & &  \\ \hline
2019 & 5.12M & & \\ \hline
\end{tabular}
}

\eat{
Finally, note that our first hypothesis on the sources of redundancy was that the download process is erroneous. However, the distinct urls from which we downloaded notebooks are 89.7\% of the overall urls. The remaining 10.3\% include both exact re-downloads but also re-downloads due to changes in the size of the notebooks (the download process searches notebooks based on their size in increasing steps of size; hence, notebooks for which their size increased during the download process may get re-downloaded). 
}

\eat{
\kk{My immediate reaction to the language results is that "oh so this is a Python corpora, not a DS corpora." Maybe we should add some explanation here that indeed notebooks are mostly for Python, but at the same time show statistics that Python is *the* DS language today.}
\fp{explained in languages}
}

\eat{
\kk{Maybe call X axis "Number of top-contributing users"?}
\fp{these are not just top-contributing, however. these are all users.}
}

\eat{
\kk{It's interesting that on average the number of notebooks per user has not increased between 17 and 19 (judging by increase in notebooks and in distinct users to be both roughly 4\%).}

\yz{Maybe a distribution of number of notebooks per user can also be helpful to draw the above conclusion?}

\SK{If it's easy to plot, can you check as feels to me as well that should be easier to read}

\fp{This is correct. avg in 2017: 12.28, avg in 2019: 12.78}
\fp{plotted the distribution. we do not have that much space. avg and std dev is fine, I think, for now.}
}

\eat{
\SK{Is it Ok to use Microsoft as example?}
\fp{removed}
}

\eat{
\kk{What are "overall parsed tree nodes"?}
\fp{addressed}
}

\eat{
\kk{Is (completely) linear our own definition? Citation needed, if not.}
\fp{addressed}
}

\eat{
\SK{can the insight on conditionals also be a WAG?}
\fp{I rewrote the text to avoid a wag.}
}
\section{Import Analysis}
\label{s:imports}

Recall that our goal with this analysis is twofold: inform practitioners of emerging common practices (e.g., pick a commonly used data preprocessing library) and assist system builders to focus their investments to better support users (e.g., optimize systems for more prominent use cases). To this purpose, after the broad statistical characterization of \autoref{s:coarsestats} we turn our attention to which libraries are used more prominently and in which combination in Python notebooks. This implicitly helps us characterize what people do in these notebooks (e.g., visualization, statistical analysis, classical ML, DNNs). We look at this through the lens of import statements (i.e., \code{import...} or \code{from...import...}). 

We begin by discussing the volume of imports and libraries across notebooks (\autoref{ss:imports:volume}). Then, we present an analysis of frequencies of imports per library (\autoref{ss:imports:frequencies}), followed by an analysis of statistical correlations (positive and negative) between library imports (\autoref{ss:imports:correlations}). We conclude the section with a coverage analysis of libraries on notebooks (\autoref{ss:imports:coverage}).

\eat{
6,978,997 total imports \ghold
2,696,629 of type from ...  \ghold

33,883,762 total 
12,839,670

41,362 libraries in \ghold
116,430 in \ghnew
}

\subsection{Landscape}
\label{ss:imports:volume}

\eat{
	We start by discussing coarse-grained statistics of the volume of imports and libraries to provide a better understanding of the landscape of the analysis in this section.
}
\autoref{tab:imports} reports the total number of imports and the fraction based on \code{from...import...} code structures for the \gh datasets. The growth of imports at $4.85\times$ is outpacing the growth of notebooks and code cells we observed in \autoref{tab:stats}. 
Also note that after deduplication, per our discussion in \autoref{s:coarsestats}, there is a reduction of 27.3\% and 31.5\% in the number of imports in \ghold and \ghnew, respectively.
Interestingly we observe a large number of unique libraries, over 116K in \ghnew (up $2.8\times$ from \ghold). \emph{This indicates the field is still substantially expanding.}

\begin{table}[h!]
	\scalebox{0.85}{
	\begin{tabular}{|ll|r|r|r|}
	\hline 
	{\bf Dataset} & & {\bf \ghold} & {\bf \ghnew} & {\bf Change} \\ 
	\hline 
	\hline
	Imports & total & 7M & 34M & $4.85\times$ \\ 
	\hline 
	 & (from) & 2.7M (38.6\%) & 13M (37.8\%) & $4.2\times$ \\ 
	 \hline 
	 Libraries & unique & 41.3K & 116.4K & $2.8\times$ \\
	 \hline
	\end{tabular}  \\ 
	}
	\caption{Import statistics for \gh.}
	\label{tab:imports}
	\vspace{-4mm}
\end{table}

\subsection{Important Libraries}
\label{ss:imports:frequencies}

Having presented the landscape in aggregate terms, we now focus our analysis on identifying what we informally refer to ``important'' libraries (i.e., libraries that have high import or usage frequencies). 

\begin{figure}
	\centering
	\includegraphics[width=0.8\columnwidth]{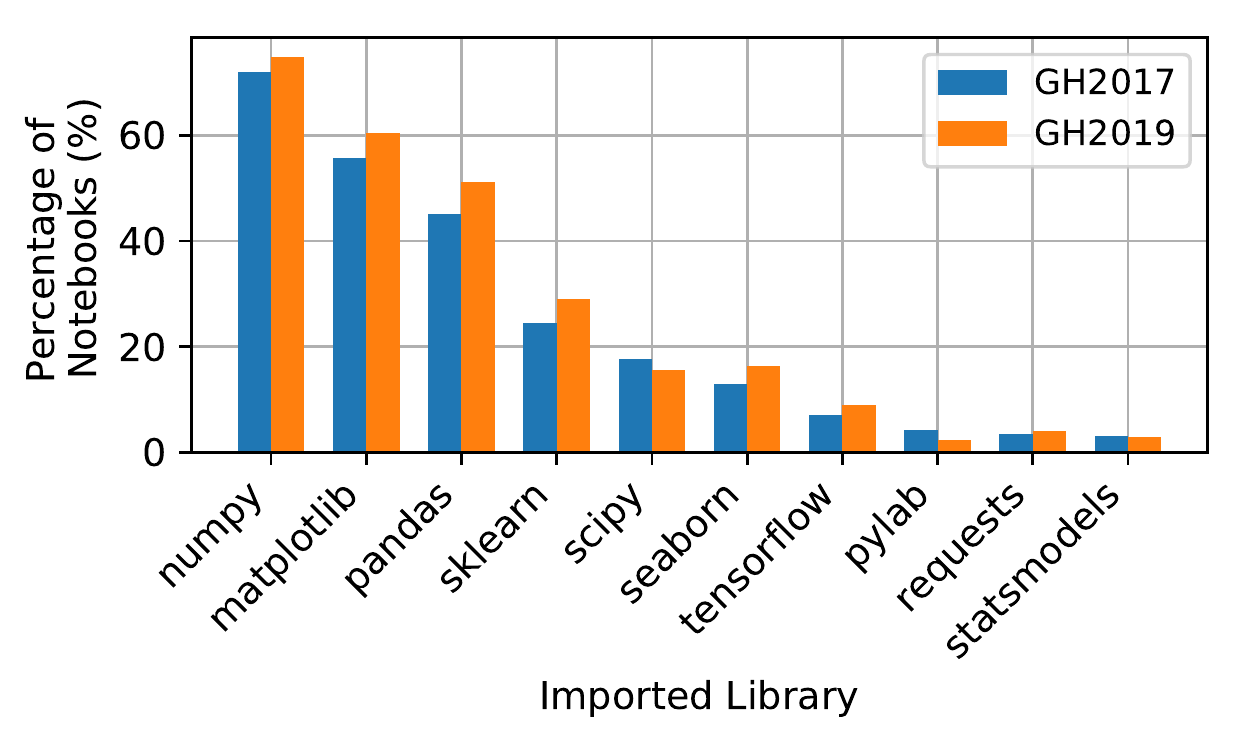}
	\vspace{-4mm}
	\caption{Top-10 used libraries across notebooks.} \label{fig:pct_pkg}
	\vspace{-4mm}
\end{figure}
\textbf{Most used libraries.} As our first metric, we consider the percentage of notebooks that import them at least once. \autoref{fig:pct_pkg}, shows the top-10 libraries in that metric. We make few observations. First we confirm a key assumption of our study: {\em notebooks are used primarily for DS activities}---the top-10 libraries imported by frequency focus on common data science tasks (i.e., downloading and communicating with external sources, processing data, machine learning modeling, exploring datasets and explaining results through visualizations, and performing scientific computations). Second, we verify our intuition that \numpy, \matplotlib, \pandas, and \sklearn are quite popular (each being imported directly\footnote{More imports could happen indirectly, e.g., \pandas internally uses \numpy, so what we provide is a lower bound of usage.}). However, these frequencies exceed our expectations.  
Third, by comparing the usage between \ghold and \ghnew, it is  really interesting to see that ``big'' (i.e., most used) libraries are becoming ``bigger'', while several libraries have lost in popularity by means of relative usage (e.g., \scipy and \pylab, as shown in \autoref{fig:pct_pkg}). This indicate a consolidation around a core of well maintained libraries. 

\wag{These results suggest that systems builders can focus optimization efforts on few core libraries (such as \numpy), but must also provide pluggable mechanisms to support a growing tail of less frequently used libraries.}

\begin{figure}   
	\centering
	\begin{subfigure}{0.23\textwidth}
		\centering
		\includegraphics[width=\textwidth]{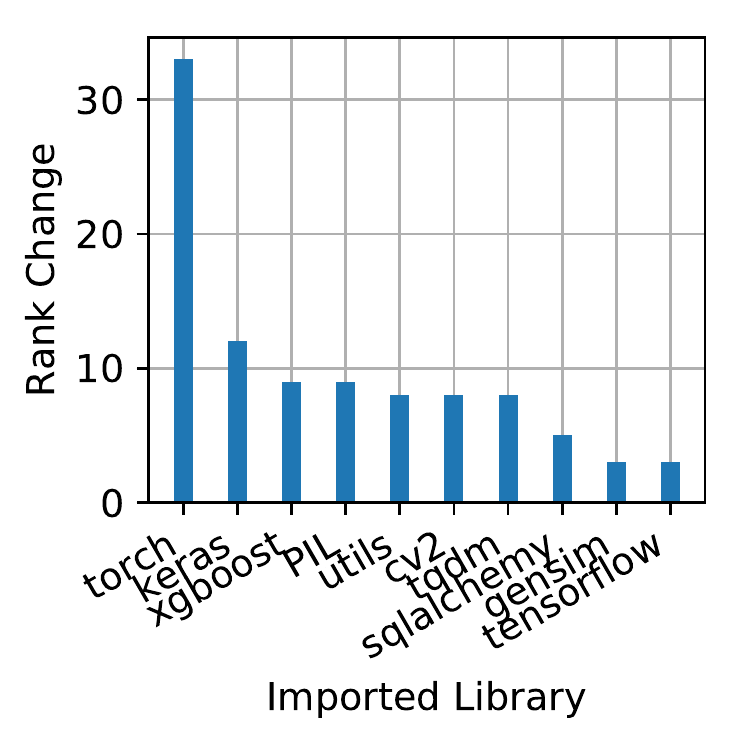}
		\caption{Rank change}
        \label{fig:rank_change}
	\end{subfigure}
	\begin{subfigure}{0.23\textwidth}
		\centering
		\includegraphics[width=\textwidth]{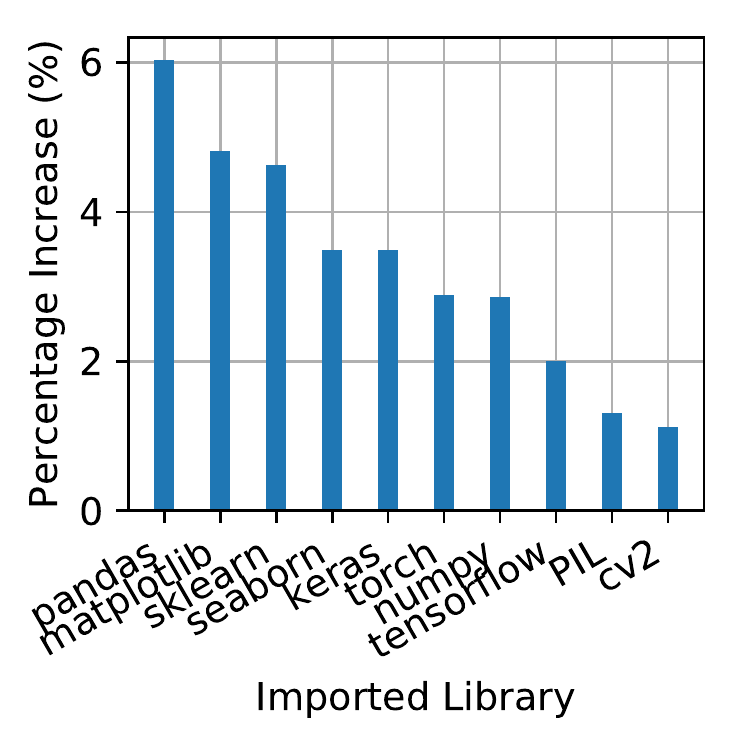}
		\caption{Percentage change}
		\label{fig:pct_change}
	\end{subfigure}
 	\caption{Top-10 libraries with most increased usage}
\end{figure}

\vspace{2mm}
\textbf{Highest ranking differentials in usage.} Comparing \ghold to \ghnew, the ranking in terms of usage changed dramatically for a few libraries. \autoref{fig:rank_change} shows the top-10 libraries that increased their usage ranking\footnote{All ranking differentials are statistically significant per Student's t-test~\cite{ttest}.} the most over the last two years\footnote{Ranking differentials are computed among the top-100 used libraries, as this is more meaningful given the larger baseline.}. Our first observation is that \torch increased the most (33 positions in our ranking) followed by \keras and \xgboost. These results confirm the  overall increased interest in deep learning and gradient boosting. Furthermore, the ranking differentials for \PIL and \cv suggest an increased interest in image processing, while the increased rank of \gensim indicates an interest for text processing. Finally, the increase of \tqdm indicates a growing interest for showing progress bars (which, in turn, indicates long-running computations or file downloading), while the increase for \sqlalchemy suggests a need to efficiently access data stored in databases. 

\textbf{Most increased in usage.} We complement the ranking differential analysis with a percentage increase in absolute terms. The top-10 libraries by highest percentage increase\footnote{The increases are statistical significant based on  Student's t-test~\cite{ttest}.} are shown in \autoref{fig:pct_change}. This shows that ``big'' libraries are getting ``bigger'' at a faster rate than average. We observe a similar pattern for  libraries related to deep learning (e.g., \keras, \torch, and \tensorflow). For this reason, we next dive deeper into a comparison of usage among deep learning libraries.

\begin{figure}[t]
	\centering
	\includegraphics[width=0.8\columnwidth]{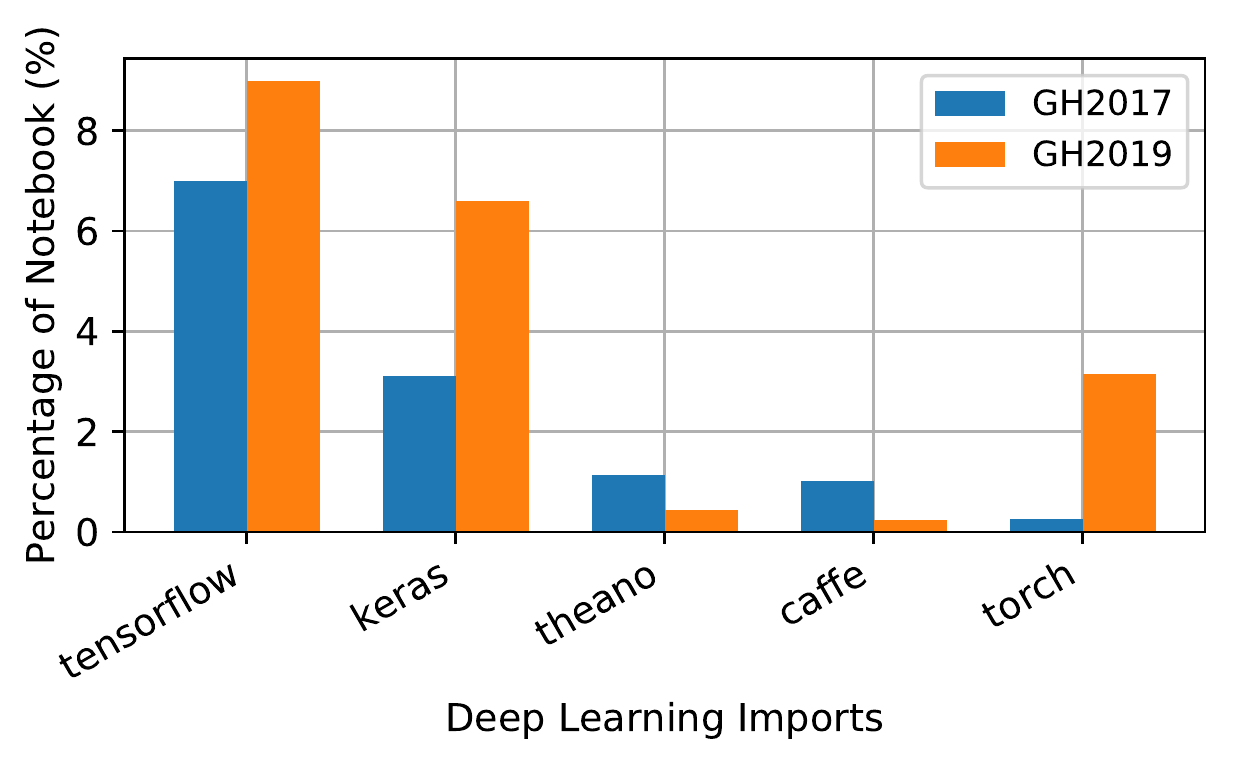}
	\vspace{-2mm}
	\caption{Deep learning libraries}\label{fig:dnn}
	\vspace{-5mm}
\end{figure}

\textbf{Comparison of usage among deep learning libraries.} \autoref{fig:dnn} shows the percentage of notebooks that use \tensorflow, \keras, \theano, \caffe, and \torch---which were the top-5 used libraries with focus on deep learning in both \ghold and \ghnew. We make two observations. First, \tensorflow, \keras, and \torch have drastically increased their usage rate (with \torch having the highest increase). Second, for both \theano and \caffe the usage rates have dropped considerably. Note that for both observations, the changes in percentages are of statistical significance based on Student's t-test.

\wag{Deep Learning is becoming more popular, yet accounts for less than 20\% of DS today.}

\begin{figure}[t!]
	\centering
		\includegraphics[width=0.8\columnwidth]{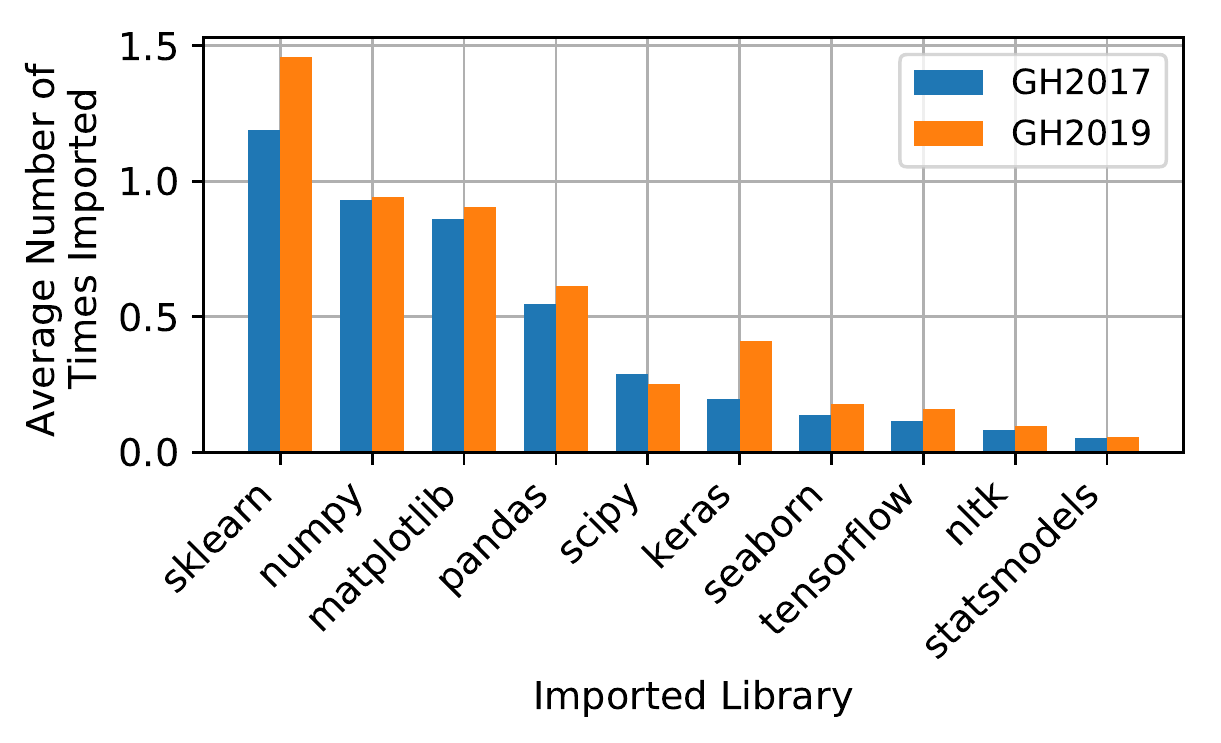}
		\vspace{-4mm}
		\caption{Average number of times to be imported}\label{fig:count_pkg}
		\vspace{-6mm}
\end{figure}
\textbf{Most imported libraries and coding patterns.} We conclude our analysis on important libraries by ranking libraries based on their import rate and the coding patterns associated with imports. More specifically, the metric we used so far considers usage of a library in a notebook if the library is used in at least one import of the notebook. Another metric, that also helps us reveal coding patterns, is the total number of times a library is involved in imports of a notebook.  In this direction, \autoref{fig:count_pkg} shows the average number of times a library is imported per notebook. We make two observations by comparing \autoref{fig:count_pkg} and \autoref{fig:pct_pkg}. First, there is a change in the overall ranking of libraries: the top-4 libraries (i.e., \sklearn, \numpy, \matplotlib and \pandas) getting rearranged, new libraries appearing in the top-10 of \autoref{fig:count_pkg} (i.e., \keras and \nltk), some other libraries changing position (e.g., \tensorflow and \seaborn), and others getting out of the initial top-10 (i.e., \pylab, \requests, and \statsmodels). The main reason for these results are due to coding patterns of data scientists in importing and using libraries. For instance, \sklearn users may prefer to import its submodules and operators explicitly, using multiple import statements. In contrast, \numpy users may use its submodules and operators directly in the code (e.g., \code{numpy.*}), and import \numpy just once.

\subsection{Correlation}
\label{ss:imports:correlations}

An interesting statistic is the co-occurrence (or correlation) of libraries in practice. 
This could indicate practitioners the need for expertise in certain combinations of libraries, and for a system builder which libraries to co-optimize. We present positive and negative Pearson correlations among libraries~\cite{pearson1896vii}---we focus on the \ghnew dataset for this analysis.

\begin{figure}   
	\centering
	\begin{subfigure}{0.23\textwidth}
		\centering
		\includegraphics[width=\textwidth]{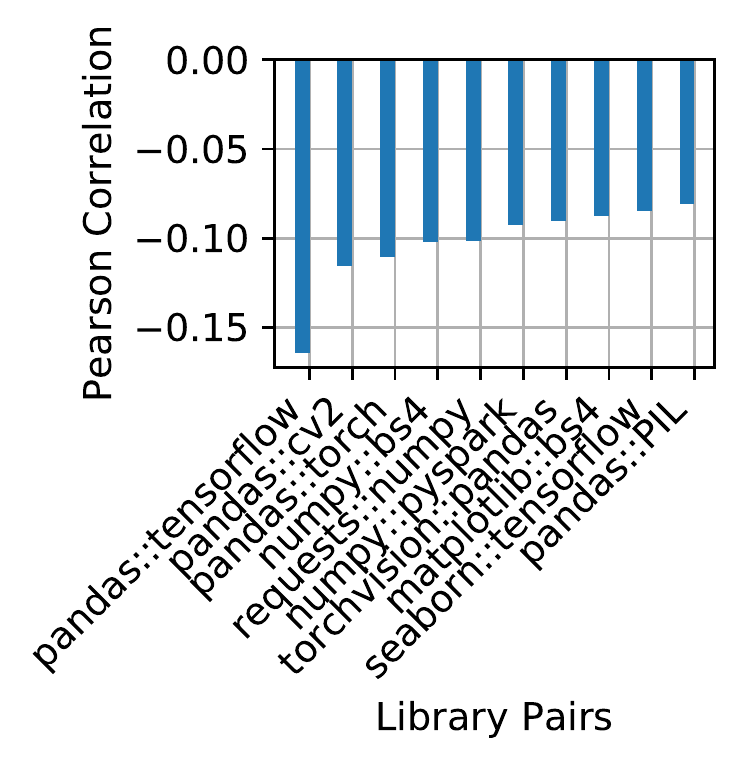}
		\vspace{-2em}
		\caption{Negatively correlated}
	\label{fig:corr_neg_correlated}
	\end{subfigure}
	\begin{subfigure}{0.23\textwidth}
		\centering
		\includegraphics[width=\textwidth]{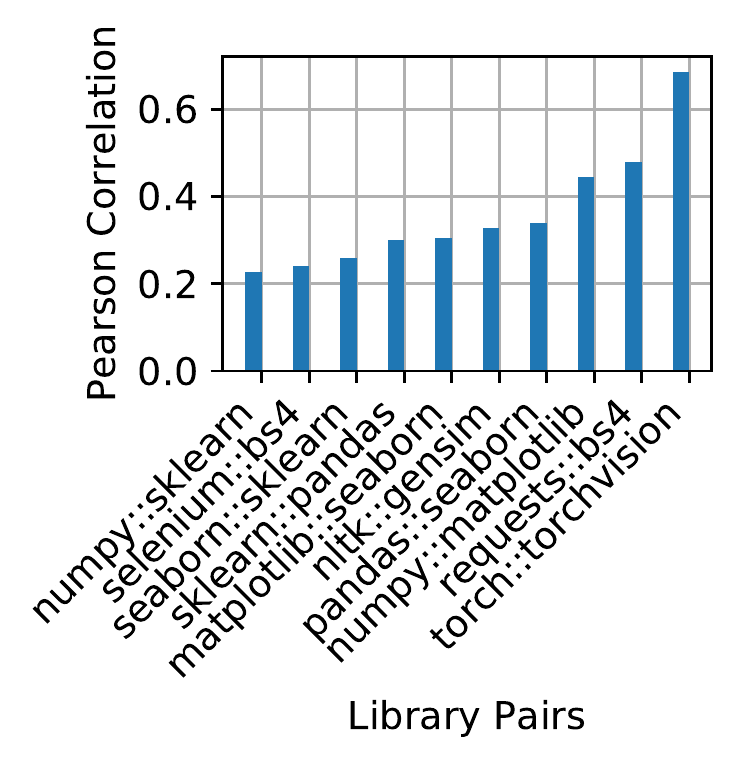}
		\vspace{-2em}
		\caption{Positively correlated}
		\label{fig:corr_pos_correlated}
	\end{subfigure}
	\caption{Top-10 correlated library pairs}
\end{figure}

\textbf{Negative Correlations.} Regarding negatively correlated libraries, \autoref{fig:corr_neg_correlated} projects the top-10 negatively correlated library pairs. We make four main observations. First, \pandas, a commonly used library for processing of tabular data, is anti-correlated with neural network frameworks (i.e., \tensorflow and \torch)---this is due to typically different data types (images vs. tabular) and support for pandas-like functionalities within neural network frameworks. This aligns with our second observation: \pandas is anti-correlated with image processing frameworks (i.e., \cv and \PIL). Our third observation is similar in nature, as we see \tensorflow and \seaborn being anti-correlated, likely because tensorflow carries its own visualization facilities. Finally, \bs, a web-page information extraction library, is anti-corraleated with \numpy. This hints at a negative correlation between array manipulation and processing of web pages, this is further confirmed by the negative correlation between \numpy and \requests.

\textbf{Positive Correlations.} \autoref{fig:corr_pos_correlated} shows the top-10 positively correlated library pairs. 
This charts provides evidence backing up common wisdom: for instance, \requests and \bs are expected to be highly correlated (i.e., \requests allows users to download urls whereas \bs allows users to extract information from web pages). Through the lenses of large corpora, such as the \gh ones we use here, we can test this hypothesis and quantify the confidence for accepting it. In the same direction, \nltk and \gensim are commonly used together for processing text corpora and building language models over them, and their correlation is reflected in \autoref{fig:corr_pos_correlated}. Furthermore, \sklearn is commonly used together with \numpy and \pandas because the input to transformers and learners of \sklearn are typically \pandas dataframes and \numpy arrays, and this hypothesis is accepted per the correlation in \autoref{fig:corr_pos_correlated}. Similar reasons of correlations exist for the other positive correlations in \autoref{fig:corr_pos_correlated} and we omit further details here.

\subsection{Coverage}
\label{ss:imports:coverage}

We conclude this section with an coverage analysis of libraries on notebooks---i.e., if we only were to support K libraries how many notebooks would be supported. We include all libraries used by \ghold and \ghnew notebooks in this analysis. 

\begin{figure}[t!]
	\centering
	\includegraphics[width=0.8\columnwidth]{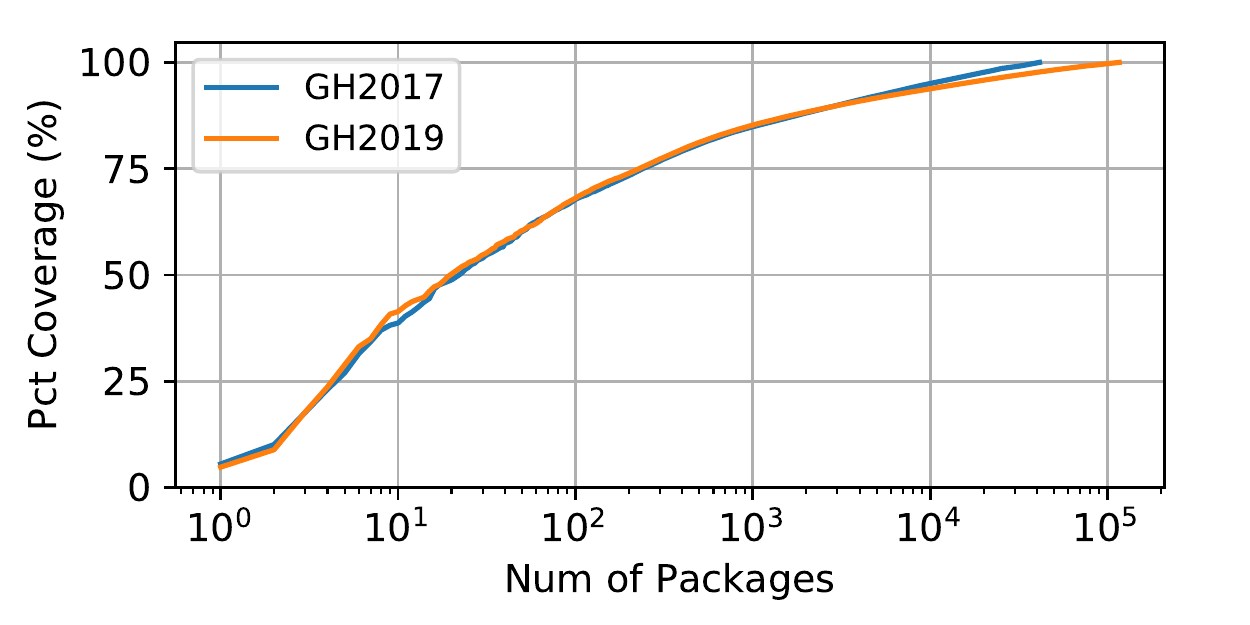}
	\vspace{-3mm}
	\caption{Percentage of notebooks to be covered}\label{fig:coverage_all}
	\vspace{-6mm}
\end{figure}

\autoref{fig:coverage_all} shows the cumulative percentage of notebooks covered (y-axis) while varying the number of libraries (x-axis). As a simple heuristic, we sort libraries from the most to the least used and pick a prefix of size K. Our main observation is that by including just the top-10 most used libraries (i.e., the ones shown in \autoref{fig:pct_pkg}), we can reach a coverage of \mysim40\% in both \ghnew and \ghold, while a coverage of 75\% can be achieved through including the top-100 most used libraries. The increase in coverage, however, is diminishing as less used libraries are added in. More interestingly, a coverage of 100\% is much harder to achieve in \ghnew than in \ghold. This further confirms that the DS field is expanding.

\eat{
In order to gain an insight on how the libraries are being used holistically, we look at the coverage percentage. The coverage is estimated by examining the percentage of notebooks that can be successfully executed by adding one more library in the same order of their ranking as in \autoref{fig:count_pkg}. In \autoref{fig:count_pkg}, for instance, the coverage to add \pandas is 14.1\%, which means 14.1\% of the notebooks use only \sklearn, \numpy, \matplotlib and \pandas. If a user installs those 4 libraries, 14.1\% of the notebooks will be runnable.

\SK{another potential WAG candidate}
}

\eat{
We successfully extracted the ``import'' information from 863,228 notebooks from 2017 and 4,059,441 notebooks from 2019. Libraries imported using \texttt{from $<$library$>$ import $<$module$>$} and \texttt{import $<$library$>$.$<$module$>$.$<$submodule$>$} were included. This section presents the statistics about the frequency of each library/module, their correlation, and the coverage, i.e. the percentage of notebooks that use only a subset of the most popular libraries.
}

\eat{
This information is extracted from the import line, and we only look at the library level, not the module/submodule level. Built-in libraries, such as \os and \sys are excluded. The top 5 are \numpy, \matplotlib, \pandas, \sklearn and \scipy for both years. It is not surprising that people use \numpy to support large, multi-dimensional data arrays, \matplotlib for data visualization, and \pandas to load/process structured data. In \autoref{s:modules}, we will look at for each library, the most popular module/function being called.
}

\eat{
\scipy ranked fifth in 2017, however, it is surpassed by \keras by a large margin in 2019.
}

\eat{
\kk{1 Structural -$>$ structured?}
\fp{1.A fixed}
}

\eat{
\kk{Can't we deduplicate to avoid this problem? I mean if numpy is "imported" three times in a notebook, we can count it once, right?}

\yz{Yes~ this would be Figure. 2?}

\fp{The deuplicated one is in figure 2, indeed. I will rephrase the text to avoid posing it as a problem. It is not a problem, just a different way to show importance.}
}

\eat{
\kk{Q. What is student t-test?}
\fp{A. Added citation} 
}

\eat{
\kk{I think we can explain here that DNN frameworks dont require pandas because they don't need data processing/featurization. I think there was a second reason Markus had mentioned, but I dont remember what that was.}

\fp{Added in negative correlations}
}

\eat{
\pandas and \tensorflow are negatively correlated (see \autoref{fig:corr_neg_correlated}), which indicates that \tensorflow users are more likely to use dataset without loading the input data as DataFrame, similarly for \torch and \PIL. For the most correlated library pairs, we have \sklearn and \pandas, \nltk and \re which are commonly used for language processing and regular expressions.
}

\eat{
covered
\fp{I stopped here my writing pass, waiting for an update on the coverage charts.} 

\fp{@Yiwen the coverage charts start with sklearn which implies that
the coverage is computed based on the ranking of Figure 5. The correct coverage needs to be computed
based on the ones of Figure 2, and the coverage needs to be computed on notebooks by modelling each notebook by the unique libraries that it has imported, per what we have discussed. Can you please update the charts with the correct computations?}
}

\eat{
\kk{Would cell coverage also make sense here? More to show popularity, rather than if a notebook is runnable. This will tell us if a library is used just once in a whole notebook or is heavily used.}
\fp{agree. just too many things to show. lets finish the first iteration, with the most important ones.}
}

\eat{
\yz{0. motivation needed?}
\fp{0.A 2nd paragraph aims to address this.}

\kk{1. High level comment for this section is that I was expecting more observations rather than simply describe what we found.}
\fp{1.A Yeah, this is important. on it.}

\kk{2. Connect this with the previous section. What is the goal here? "Having seen some general statistics about the notebooks and cells, we now focus on what practitioners are doing with these notebooks by analyzing the libraries they use". I guess these are only for Python, so we should state it. Also the percentage of all notebooks would be nice to give instead of only absolute numbers.}

\fp{2.A Handled in introductory paragraph}

\kk{3. Coverage is not very clear here. You mean the percentage of notebooks that exclusively use these only these popular libraries? Currently it reads as the percentage of notebooks that use only a subset instead of all of the most popular libraries.}

\fp{3.A I removed that intro paragraph.}

\kk{4. Was there something challenging in collecting these imports? I think you are alluding to the fact that there are multiple ways to import something, but it does not come across.}

\fp{4.A. There was nothing hard about collecting these imports. Just a 70loc extractor. Tried to removed text indicating complexity of this extraction process.}

\kk{5. I'd drop "successfully" unless you explain why the rest was not successful or why it was challenging.}

\fp{5.A I dropped this paragraph.}
}

\eat{
\SK {shouldn't we have a WAG (ideally linked all the way from intro) on the big getting bigger?}

\fp{Imho, bigs getting bigger is not a guess, for we show it from our analysis. A guess could be that bigs are getting bigger in general, but I do not believe I am saying this here. do I?  I am leaving this discussion open for Carlo and Markus to see in case they find it interesting for the intro.}
}

\section{Pipelines}
\label{s:pipelines}

Our analysis so far has focused on understanding data science projects based on the libraries they are using (\autoref{s:imports}). In this section, we dive deeper into well-structured  pipelines (namely, \sklearn and \tlc pipelines) to provide an ever finer-grained view of the data science logic, that is also optimizeable and manageable~\cite{cidrvision,schelter2017automatically}. More specifically, we start by providing volume statistics of pipelines in \gh and \tlc to get a better view of the landscape (\autoref{ss:pipelines:landscape}). Then, we focus on the length of pipelines as a measure of their complexity (\autoref{ss:pipelines:length}). Next, we provide an overview of the number of operators used in pipelines, and we make the case for the need of further functionality (\autoref{ss:pipelines:ops}). Furthermore, we present an analysis of frequencies of learners and transformers in pipelines to point out common practices (\autoref{ss:pipelines:learntransform}). We conclude this section with a coverage analysis of operator on pipelines to better understand the complexity of supporting pipeline specifications in systems for ML (\autoref{ss:pipelines:coverage}).

\subsection{Landscape}
\label{ss:pipelines:landscape}

We start by providing a description of pipelines that we use in the analysis of this section along with their volume.

\textbf{Description of \sklearn pipelines.} To understand the complexity of well-formed pipelines in the \gh corpora, we focus on the \sklearn pipelines primarily because (a) of  their popularity; (b) multiple systems for ML aim to manage and optimize them~\cite{cidrvision,schelter2017automatically}; and (c) their specification resembles the one of \tlc, enabling us to compare public with enterprise-grade pipelines. Specification-wise, \sklearn provides \code{Pipeline} and \code{make\_pipeline} constructs for generating pipelines.  Such pipelines consist of a sequence of transformers (e.g., \code{StandardScaler}) and learners (e.g., \code{LogisticRegression}). (Intermediate steps of each pipeline are transformers and the last step can be either a learner or a transformer). Hence, their main purpose is to transform input data for either training or inference purposes. Finally, note that pipelines can incorporate sub-pipelines through the use of \code{FeatureUnion} and \code{make\_union} constructs that concatenate the results of multiple pipelines together.

\textbf{Volume of \sklearn pipelines.} We now compare the volume of \sklearn pipelines between \ghold and \ghnew. From \ghold, we managed to extract only 10.5k pipelines. The relatively low frequency (with respect to the number of notebooks using \sklearn discussed in \autoref{s:imports}) indicates a non-wide adoption of this specification. However, the number of pipelines in the \ghnew corpus is 132k pipelines (i.e., an increase of $13\times$ or an average of 181.5 pipelines getting committed daily on \gh since 2017). As such, we believe that the ``declarative'' specification of data science logic, that opens up optimization and management opportunities, is gaining in momentum. We note, however, that an interesting future work in this space is to compare the explicitly specified \sklearn pipelines with the ones specified implicitly (i.e., using \sklearn operations and tightening them together imperatively).

\textbf{Description of \tlc.} \tlc API is similar to \sklearn: operators are assembled into a data flow graph. Each operator is either a transformer or a learner. \tlc is much richer in ``data-massaging'' operators than \sklearn, as it was developed with the goal of capturing end-to-end pipelines from domain object to domain decision. Users of \tlc can author their pipelines in several ways: through a typed imperative language, in Python through bindings, or using \tlc's scripting language. For our analysis here we use telemetry data that records the scripting API usage.

\textbf{Volume of \tlc pipelines.} Starting from the 88M telemetry events we extracted 29.7M pipelines from 2015 onwards. We found that many pipelines use the exact same set of operators (albeit with different parameters). This is because \tlc provides suggested defaults for many DS tasks as well as built-in parameter sweeps. We kept only one copy of these pipelines, which left us with 2M unique pipelines. We analyze those in this section.

\eat{
\fp{Backing up numbers before rounding}
29,737,958
110,697,560 
1,947,504 
}

\eat{
out of 210k notebooks that use \sklearn, we found that only 4.6k of them contain \sklearn pipelines. In aggregate, 
}

\eat{
\kk{As a non-TLC expert, not sure I see the different with scikit-learn pipelines here. I understand that, unlike ONNX, TLC has more transformers, but compared to a scikit-learn pipeline, whether the data preprocessing are part of the ML library or not, does not seem different. Might still be worth mentioning in that TLC pipelines do not have to rely to other external libraries (which has advantages and disadvantages--better control of what you use, but can't take the innovation of systems dedicated to preprocessing).}
\mi{I hope that the text is clearer now. \tlc does support external lib e.g., tensorflow and eventually soon torchsharp.}
\kk{TBH, I don't think the description adds something here, especially given we don't have too much space. We explain that AnonSys has in some sense more native support for some pipeline operators, but it does not seem that this detail changes the analysis or contributes to its understanding. If it does, then we should keep it and make more clear why it does. Otherwise, I'd just keep the first and maybe last period from this para and fold them to the following para.}

\fp{ I removed those two sentences: 1. ``(i.e., \pandas and \sklearn functionalities are merged into \tlc)''  2. ``After training \tlc models can be end-to-end serialized and deployed directly into applications, such that inference and training use the same code path~\cite{rule-of-ml}''.  For 1. sklearn can take input from numpy and other sources. For 2. it does not add something to the analysis. Also, \sklearn pipelines can be pasted in mlflow these days or managed by sagemaker. I left the original text with the 2 unremoved.}
}

\eat{
\textbf{Description of \tlc.} \tlc API follows the \sklearn design: pipelines are graph of operators whereby each operator can be either a transformer or a learner. In contrast to \sklearn, however, \tlc operators incorporate ``data-massaging'' operators such that pipelines can express end-to-end models (i.e., from data ingestion to training and prediction). This characteristic simplifies model deployment: after training \tlc models can be end-to-end serialized and deployed directly into applications, such that inference and training use the same code path~\cite{rule-of-ml}.
Users of \tlc can author they pipelines in several ways: through a typed imperative language, through Python bindings, or using \tlc's scripting language. For the analysis in this section we use the telemetry information recording \tlc scripting API usage.
}

\eat{
\textbf{Volume of \tlc pipelines.} We started with an initial sample of around 88M telemetry events. Among those we loaded into the database 29.7M pipelines used for training models. Within those pipelines, we observe that a total of 110.6M operators are instantiated. Among all pipelines, we count 2M unique pipelines using 698 unique operators. (Note that the total number of \tlc operators is 210. The high number of unique operators is due to typos and telemetry parsing errors.) We will specifically use unique pipelines to drive the following analysis.
}

\eat{
removed the discussion on pipeline operators to move it to S5.2
Within those pipelines, we observe that a total of 110.6M operators are instantiated. Among all pipelines, we count 2M unique pipelines using 698 unique operators. (Note that the total number of \tlc operators is 210. The high number of unique operators is due to typos and telemetry parsing errors.) We will specifically use unique pipelines to drive the following analysis.
}

\eat{It is supported by a lot of AI platforms, such as Google Cloud, etc. }

\eat{For 2017, among the 210,765 notebooks that use \texttt{Sklearn}, we extracted 10,500 \texttt{Pipeline} from 4,681 notebooks. {\color{blue} it is a relatively small portion but might represent more advanced users or pipelines that with less complex operations.}}

\eat{
\kk{Why did we pick scikit-learn pipelines? I guess because of their popularity and because that's the only known way to create pipelines. Let's say it.}

\kk{Explain what the purpose of a pipeline is. You take raw data, pass them through various transformers to prepare them, then do the training. Connect that with what you told me you see in the pipeline operators: they either include a single learner at the end, or they don't even include a learner. Say that there can be steps outside of the pipelines too, so the practical length is even longer.}
}

\eat{
\kk{Along with their increase in popularity between 17 and 19, I'd add a personal WAG that we believe in pipelines as a way to better structure the code and share across users.}
}

\eat{
\kk{Say that we scikit-learn is for non-DNNs, but also DNNs typically do not require pipelines as much, given no preprocessing is needed.}
\fp{I cannot back that up.}
}
\eat{
\kk{Are all pipelines for training? Or inference too?}
\fp{We can't tell}
}

\eat{
Overall, we believe this is an interesting slice of the \gh data because it provides well-formed pipelines that enterprise-grade systems for ML~\cite{cidrvision,schelter2017automatically} aim to manage and optimize. Also, the actual specification has many similarities with the \tlc one. Hence, it enables us to provide a comparison between public and enterprise-grade pipelines.
}

\eat{
\AF{Maybe percentages are easier to read?The 10,500 pipelines contain implicit + explicit? This needs to be clarified. The 132K pipelines in 2019 are implicit + explicit?}
\fp{clarified and tried to made it easier to read.}
}

\eat{
\AF{Overall, do we need full numbers? Maybe we can approcimate--it might be easier to read}
\fp{Changed}
}

\subsection{Pipeline Length}
\label{ss:pipelines:length}

\begin{figure}[t!]
	\centering
	\includegraphics[width=.9\columnwidth]{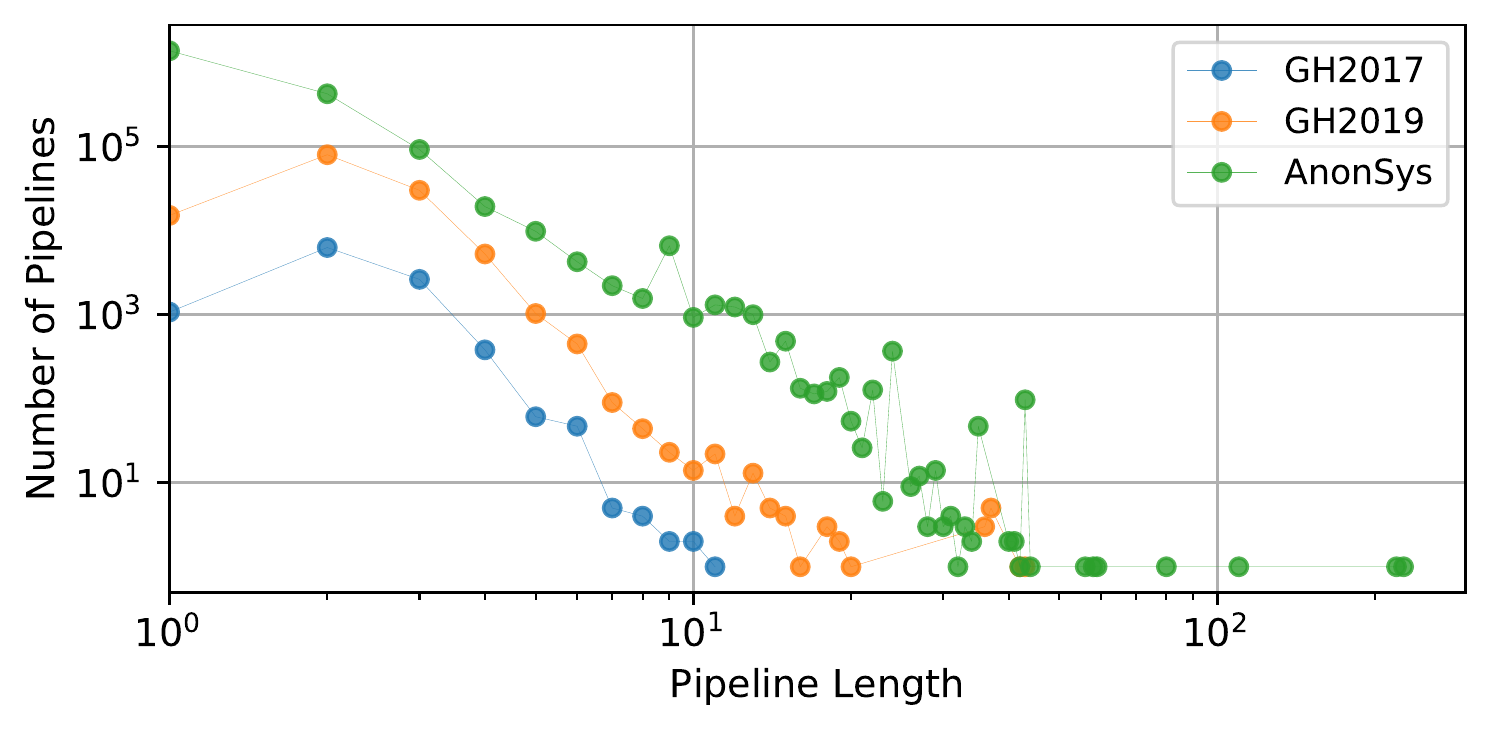}
	\vspace{-4mm}
	\caption{Number of pipelines per pipeline length.}
	\vspace{-6mm}
	\label{f:pipeline-length}
\end{figure}

As a proxy for pipeline complexity we consider their length. \autoref{f:pipeline-length} shows the \#pipelines per pipeline length for \ghold, \ghnew, and \tlc to drive our discussion. 

\textbf{\sklearn.}  We make two observations for \sklearn pipelines. First, both \ghnew and \ghold are right-skewed. Most pipelines have length of 1 to 4. Second, both corpora contain long pipelines. However, the length has increased substantially  in \ghnew. Many pipelines have length above 11 (i.e., max length in \ghold), with a max length of 43. Finally, note that in some cases  (5\% in \ghold and 5.6\% in \ghnew) the length reported here is a lower bound. In these cases, the pipeline is assembled out of multiple sub-pipelines, each held in a variable. Our current analysis treats these sub-pipelines as a single operator. We leave a taint analysis to future work to remedy this.

\textbf{\tlc.} \autoref{f:pipeline-length} also reports the number of pipelines per pipelines length for \tlc. Similarly to the \sklearn pipelines, \tlc pipelines are right-skewed. Most pipelines have a length of 1 to 4. In contrast to \sklearn, however, \tlc pipelines can get lengthier, reaching a max length of 227.

\textbf{Comparison.} \tlc pipelines tend to be longer than those in \sklearn. This can be attributed to the end-to-end nature of \tlc pipelines: They contain the full data flow from I/O through transformation to model. In \sklearn, much of the early stages of this are handled by Python code or libraries not captured in \autoref{f:pipeline-length}.
\wag{The design of \tlc points to a possible future where DS pipelines are fully declarative.}

\eat{Another reason, however, is that \sklearn pipelines are specified in Python and may take as input the result of other operations, whereas \tlc pipelines run independently. Hence, \sklearn pipelines may get inputs (e.g., dataframes or the outputs of other pipelines)  that are computed through lengthy operations (e.g., cleaning) and whose length is not reflected in \autoref{f:pipeline-length}. We believe this is an interesting space for future investigation (in association with the taint analysis we discussed above).}

\eat{
\textbf{Comparison.} \autoref{f:pipeline-length} also allows us to compare the length between \sklearn and \tlc pipelines. Our main observation is that, while both types of pipelines have a length that is right-skewed and centered around 1 and 4, \tlc pipelines can get lengthier. We believe this can be attributed to two reasons. An obvious one is that enterprise-grade pipelines are lengthier due to the introduction of complex operations as part of complicated business logic. Another reason, however, is that \sklearn pipelines are specified in Python and take as input the result of other operations, whereas \tlc pipelines run independently. Hence, \sklearn pipelines may be get as input data (e.g., dataframes or the outputs of other pipelines)  that are computed through lengthy operations (e.g., cleaning and wrangling) and whose length is not reflected in \autoref{f:pipeline-length}. We believe this is an interesting space for future exploration and investigation (in association with the taint analysis we discussed above).
}

\eat{Next, we discuss our observations on \sklearn and \tlc pipelines in isolation, followed by a comparison of the two.}

\eat{
\AF{Figure 12 caption: maybe rename it to "Number of pipelines as their length varies"}
\fp{Not a line chart per se---hence, not sure about variation.}
}

\eat{
\AF{Figure 12 x-axis: maybe rename it to Pipeline Length}
\fp{done!}
}

\eat{
\AF{Figure 12 labels: say GITHUB2019, GITHUB 2017 instead of 2017, 2019?}
\fp{done. requires update of other charts.}
}
\subsection{Operators and the need for external functionality}
\label{ss:pipelines:ops}

To better understand the complexity of Data Science across time and datasets, we consider the number of (unique) operators in this section.

\textbf{\sklearn.}  The pipelines in \ghold and \ghnew contain 25.4K and 309.2K operators in aggregate. Most operators in both datasets are specified as function calls inlined in pipelines: 72.5\% in \ghold and 75.6\% in \ghnew. 	Of those function calls, 587 and 3,397 are unique in \ghold and \ghnew, respectively. This indicates that not only do pipelines get longer from \ghold to \ghnew, they also get more diverse.

\textbf{\tlc.} We identified 110.6M total operators. 536 are unique. Focusing on operators that have a equivalents in \sklearn, we identify 463 unique operators.

\eat{
  ~\autoref{fig:cdf-tlc} shows the cumulative number of pipelines that are executable by only using the transforms and learners specified on the x-axis. As we can see, with only five operators we already support 80\% of the pipelines. This information is really insightful for practitioners because it suggests that only a handful of transforms and learners needs to be supported efficiently in order to cover the vast majority of the pipelines.
}

\textbf{Comparison.} On first glance, \sklearn pipelines have more unique operators than the \tlc ones. However, both \tlc and \sklearn and allow for user defined operators (UDO) in their pipelines. However, the two systems go about this differently: In \tlc, a UDO is wrapped by a single operator and therefore, all 23k UDOs in our dataset show up as a single operator. In \sklearn, UDOs are introduced as separate operators. This explains the large number of unique operators observed in the \gh datasets.

\textbf{The need for more functionality.} The pipelines analyzed here require functionality that is not natively supported by \tlc or \sklearn, leading users to introduce UDOs. Some of those UDOs are unavoidable due to the rich set of domains DS is applied to. However, we speculate: \wag{Data Science evolves faster than the systems supporting it. And large corpora of data science workloads, such as the ones we focused on here, can help reveal of ``what is missing?'' from current libraries.}

\eat{
  We make two interesting observations when comparing the number of operators in \sklearn and \tlc pipelines a) that \sklearn pipelines seems to have more unique operators and b) both types of pipelines involve operators other than the ones natively supported by \sklearn and \tlc. Our discussion next will help reveal both. First, note \sklearn pipelines can include any operator that complies with the transformer and learner interface of \sklearn. As such, users either construct and use their own transformers and learners in \sklearn pipelines or use ones provided by libraries either than \sklearn (e.g., \code{xgboost.XGBClassifier}). Second, note that \tlc also provides this capability through UDFs. These account for 23k operators in total in \tlc pipelines. Unfortunately, however, we do not have access to the particular UDFs and the results on \tlc above account all UDFs as 1 operator---hence, the main difference on unique operators between \tlc and \sklearn. Regardless, however, the main outcome of this analysis is our second observation: for \sklearn and \tlc pipelines there are many situations where users need functionality other than the one natively supported by \sklearn and \tlc.
}

\eat{
  \AF{Not sure I understand this sentence}
  \kk{If it means that those previous high numbers are due to variables being counted as unique, I don't like it much (at least needs to be stated clearly). The next numbers are better.}
}
\subsection{Learners and Transformers}
\label{ss:pipelines:learntransform}

In our discussion above, we focused on operators in aggregate terms. To analyze individual operators, we proceed into ranking them by frequency--which is helpful for prioritizing efforts in developing systems for ML, and exposing common practices among practitioners. We do so by first classifying operators natively supported by \sklearn and \tlc to learners and transformers:

\textbf{\sklearn.}  Top-5 transformers in \ghnew are \code{StandardScaler}, \code{CountVectorizer}, \code{TfidfTransformer}, \code{PolynomialFeatures}, \code{TfidfVectorizer} (in this order).  Same are the results for \ghold with the difference that \code{PCA} is 5\textsuperscript{th} instead of \code{TfidfVectorizer}.  Regarding learners, Top-5 in both \ghold and \ghnew are \code{LogisticRegression}, \code{MultinomialNB}, \code{SVC}, \code{LinearRegression}, and \code{RandomForestClassifier} (in this order). Analyzing the frequency of operators per position reveals that~\code{StandardScaler} dominates the first position.

\eat{(and \code{StandardScaler} more specifically).}

\eat{
  (Note that for both learners and transformers, top-10 and frequency plots are included in the Appendix.) 
}
\eat{
  The top transformers are respectively \code{OneHotEncoder}, \code{TfidfVectorizer}, \code{Imputer}, \code{CountVectorizer} (in this order), while the top-5 learners include \code{Gradient Boosting Classifier}, \code{Random Forest Classifier}, \code{SDCARegression}, \code{PoissoneRegression}, and \code{AveragedPerceptron}.
  We did the same exercise as for \sklearn, and we computed the top used transformers and learners within \tlc pipelines to validate that what DS do on notebooks is not that different that what is found in industry pipelines.
}

\textbf{\tlc.} Top transformers are \code{OneHotEncoder}, \code{TfidfVectorizer}, \code{Imputer}, \code{Tokenizer}, \code{CountVectorizer} (in this order), while the top learners include \code{Gradient Boosting}, \code{Random Forest}, \code{SDCA}, \code{PoissoneRegression}, and \code{AveragedPerceptron} (in this order).\footnote{For anonymity purposes, we report the \sklearn operators that are the closest equivalent to the ones found in \tlc.}

\textbf{Comparison.} Normalizers are not within the top-10 for \tlc, while dimensionality reduction operators (e.g., PCA) are not even in the top-30. Both are very popular in \sklearn pipelines. These differences are easily attributed: \tlc adds these automatically based on the needs of downstream operators. The tree learners in \tlc are optimized to deal with sparse, high-dimensional data, which lessens the need for dimensionality reduction.
Regarding learners, \code{Gradient Boosting} and \code{Random Forest} are more popular in \tlc than \sklearn. We believe this to be due to their relative quality and the tasks observed in \companyname.
\subsection{Coverage}
\label{ss:pipelines:coverage}

Here, we focus on the full set of operators of \tlc and \sklearn to compute their coverage on the respective set of pipelines. Note that we consider a pipeline covered if all of its operators have been supported.  Such analysis is helpful for developers of DS systems to prioritize their efforts.
\eat{
	To save space, we discuss our observations collectively, without discussing insights for \sklearn and \tlc in isolation and in comparison.
}

\begin{figure}
	\centering
	\includegraphics[width=.8\columnwidth]{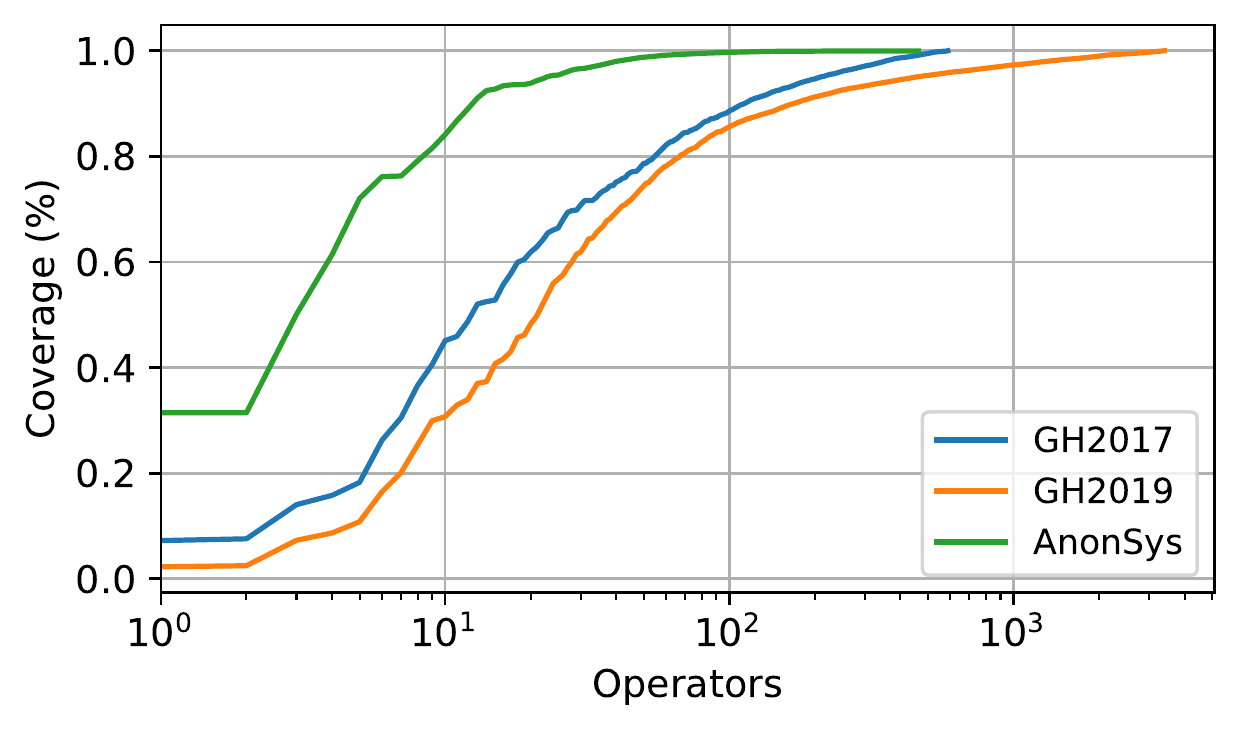}
	\vspace{-4mm}
	\caption{Coverage of operators on pipelines.}
	\label{f:cdf}
	\vspace{-6mm}
\end{figure}

\autoref{f:cdf} shows the coverage for \ghold, \ghnew, and \tlc while increasing the number of operators in the descending order of operator frequency. We make three main observations. First, the top-100 operators cover more than 80\% of pipelines across all datasets. Second, the same number of operators can cover less of \ghnew than \ghold. This is indicative of the overall increased complexity of \sklearn pipelines. Finally, the top-10 operators in \tlc cover \mysim80\% of all pipelines.
\eat{
	Recall, however from our discussion in \autoref{ss:pipelines:ops}, that UDFs are considered a single operator. In this ranking this operator is 7\textsuperscript{th}, and explains to some extent this high coverage. Still, however, even with 6 operators (i.e., without the UDF one) the coverage is ~\mysim70\%.
}

\wag{Optimizing only relatively few operators can have a tremendous impact on the overall set of pipelines.}

\eat{
Having presented our analysis on \sklearn and \tlc pipelines in isolation, we now attempt a comparison between the two. To do so, we compare the pipelines in terms of their length, the coverage of the operators on the pipelines, and on individual operators.

\textbf{Length.} \autoref{f:pipeline-length} depicts numbers of pipelines per pipeline length for all three corpora (i.e., \ghold, \ghnew, and \tlc). We make two observations. 
}

\begin{figure*}[t!]
\centering
\begin{minipage}[t]{.33\textwidth}
  \centering
  \includegraphics[trim={0.25cm 0 0.25cm 0},clip,width=.95\textwidth]{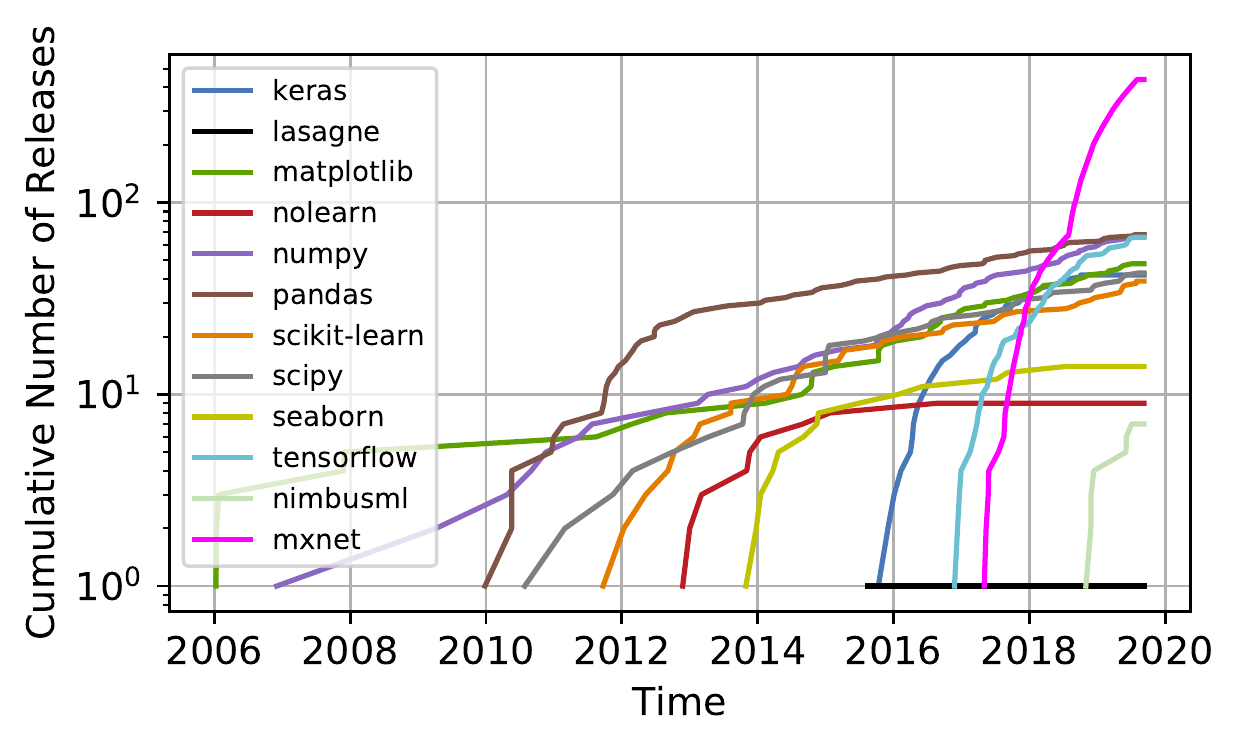}
\vspace{-1.5em}
  \captionof{figure}{Release frequency of libraries.}
 \label{fig:version1}
\end{minipage}%
\begin{minipage}[t]{.33\textwidth}
  \centering
  \includegraphics[trim={0.25cm 0 0.25cm 0},clip,width=.95\textwidth]{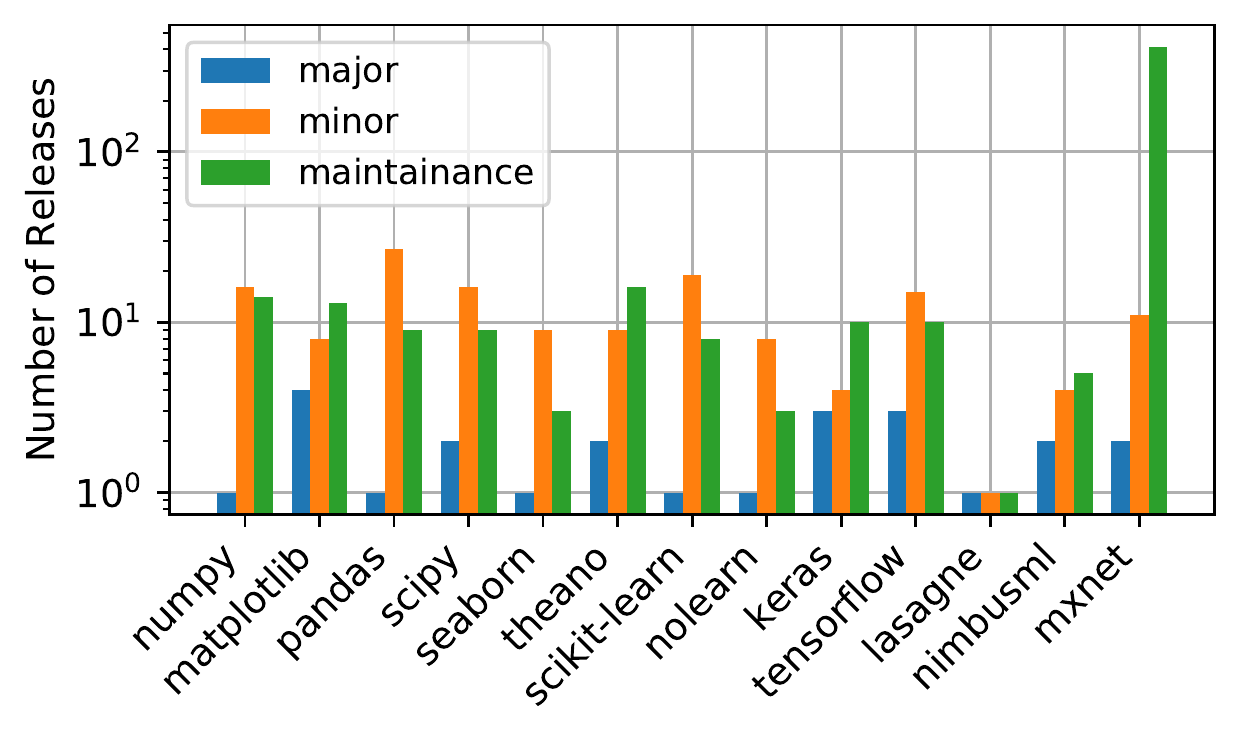}
  \vspace{-1.5em}
  \captionof{figure}{Major, minor, and maintenance \#releases per library.}
 \label{fig:version_count}
\end{minipage}%
\begin{minipage}[t]{.33\textwidth}
  \centering
  	\includegraphics[trim={0.25cm 0 0.25cm 0},clip,width=.95\textwidth]{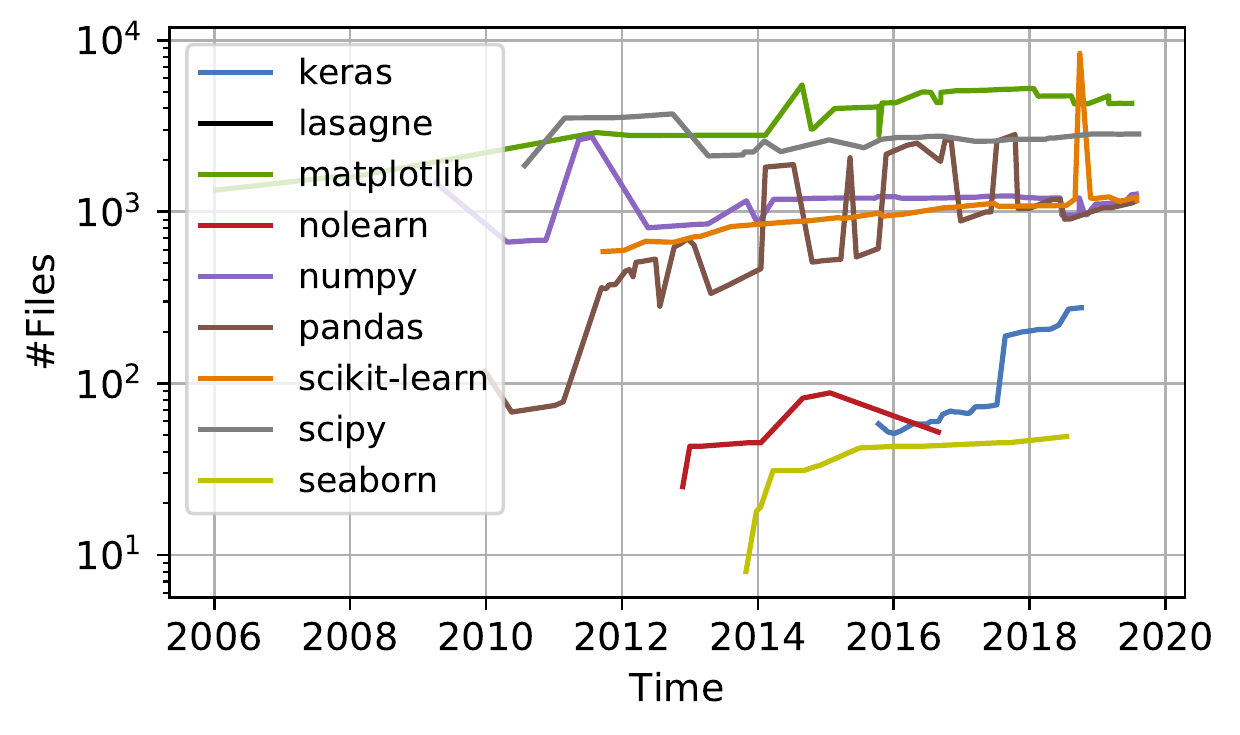}
\vspace{-1.5em}
  \captionof{figure}{\#files per library over time.}
 \label{fig:file_count}
\end{minipage}
\end{figure*}

\section{Libraries}
\label{s:modules}

We conclude our analysis by taking a closer look at individual Python libraries that data scientists rely on: \sklearn, \numpy, \matplotlib, \pandas, \scipy, \seaborn, \theano, \nolearn, \keras, \lasagne, \nimbusml, \tensorflow, and \mxnet. 
\eat{
	These libraries are a) the most used libraries in Python notebooks (as reflected by our analysis in~\Cref{s:imports}) and b) considered important for data science purposes (even if not projected as most used by our analysis).
}
Our goal with this analysis is to better understand their temporal evolution and whether they have reached a consensus on the functionality they expose (both of which can drive decision making of developers and practitioners). To do so, we first analyze their release frequencies as reflected on \pypi, followed by a more in-depth analysis on their source as provided in \pypi.

\subsection{Release Analysis}

We start with the analysis of releases by means of release frequencies over time, followed by a break down of releases to release types (i.e., major, minor, and maintenance).

\textbf{Release frequency.} \autoref{fig:version1} shows the cumulative number of releases per library over time. We make three observations. First, \matplotlib, \numpy, and \pandas are constantly being updated. Especially after 2016, releases become more frequent for \matplotlib and \numpy. Second, we observe a rapid increase in releases of packages such as \keras, \tensorflow, and \theano. This is likely driven by the current attention to the DNN area. We also observe that both "classical" ML and DNN toolkits see constant releases: \sklearn and \tensorflow are good examples of this.

\textbf{Major, minor, and maintenance releases.} 
The simple number of releases does not reflect the type of changes provided by releases. To do so, we look into the release versions. Following the version scheme defined in~\cite{pep440}, for a library release defined as X.Y.Z, we consider major to be X, minor to be X.Y, and maintenance to be X.Y.Z. \autoref{fig:version_count} shows the number of major, minor, and maintenance releases for each library. 
We make the following main observations. First, \matplotlib, \keras, and \tensorflow have the largest number of major releases. Second, we observe that many libraries did not have multiple major releases. In particular, \numpy, \pandas, \seaborn, \sklearn, and \nolearn have only one major release. Third, \pandas, \sklearn, and \tensorflow have the largest number of minor releases. Fourth, \mxnet has a much larger number of maintenance releases compared to all other libraries. 
\eat{
	We speculate that this is due to the strict release guidelines of Apache in an area that requires rapid development and engagement with the community.
}

\subsection{Source Analysis}
To better understand the code evolution in terms of files, functions, and classes we also explore the source of every release of every library.
For this analysis, we do not consider \tensorflow, \nimbusml, and \mxnet, as well as a few releases of other packages, because their \pypi package contains native code that we cannot currently analyze. Finally, note that our analysis relies on inspecting the source of releases without installing them.

\textbf{Files.} \autoref{fig:file_count} shows the number of files per library release over time. We observe that all libraries have gone through one or more phases of rapid increase in number of files.  Also, all libraries have gone through one or more phases of source file removals (e.g., due to removal of deprecated functionality and documentation).  Lastly, some libraries (e.g., \pandas) have a big variance in the number of files across releases. This is due to the first two observations, but also because some libraries maintain multiple major/minor releases at the same time. For instance, two minor releases maintained (through multiple and different maintenance releases) in parallel over time may result in high variance in this analysis if the maintenance releases have a big difference in terms of number of files. Finally, although several libraries have started stabilizing around a specific number of files, this should not be interpreted definitively as consensus in exposed functionality. Files in this analysis do not contain only the Python files that exposes functionality that data scientists use in their work, but also documentation and testing files. For this reason, we next dive deeper into the analysis of classes and functions from Python files of each library.

\begin{figure*}[t!]
    \centering
    \begin{subfigure}[t]{0.24\textwidth}
        \centering
        \includegraphics[trim={0.25cm 0 0.3cm 0},clip,width=\textwidth]{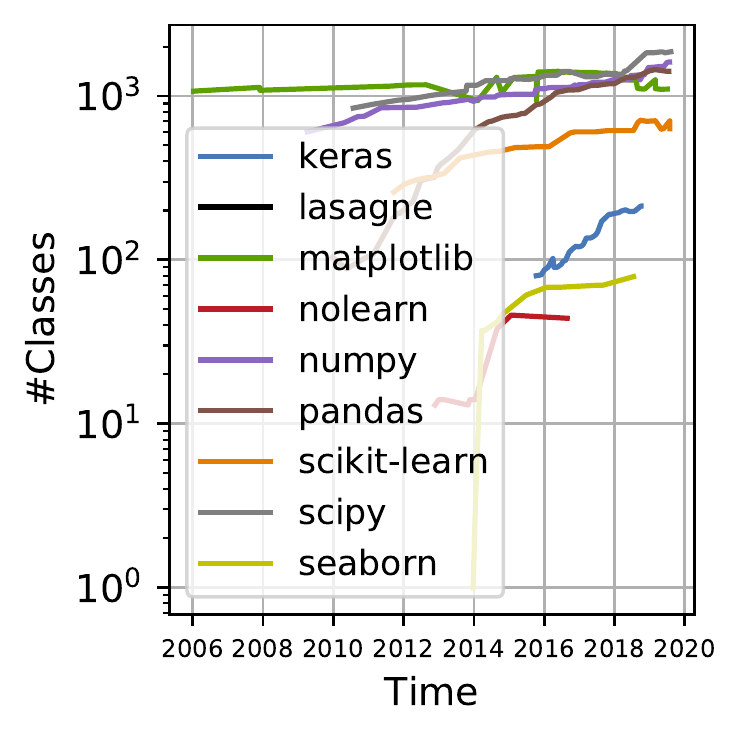}
		\vspace{-2em}
		\caption{}
        \label{fig:class_count}
    \end{subfigure}%
    ~ 
    \begin{subfigure}[t]{0.24\textwidth}
        \centering
        \includegraphics[trim={0.25cm 0 0.3cm 0},clip,width=\textwidth]{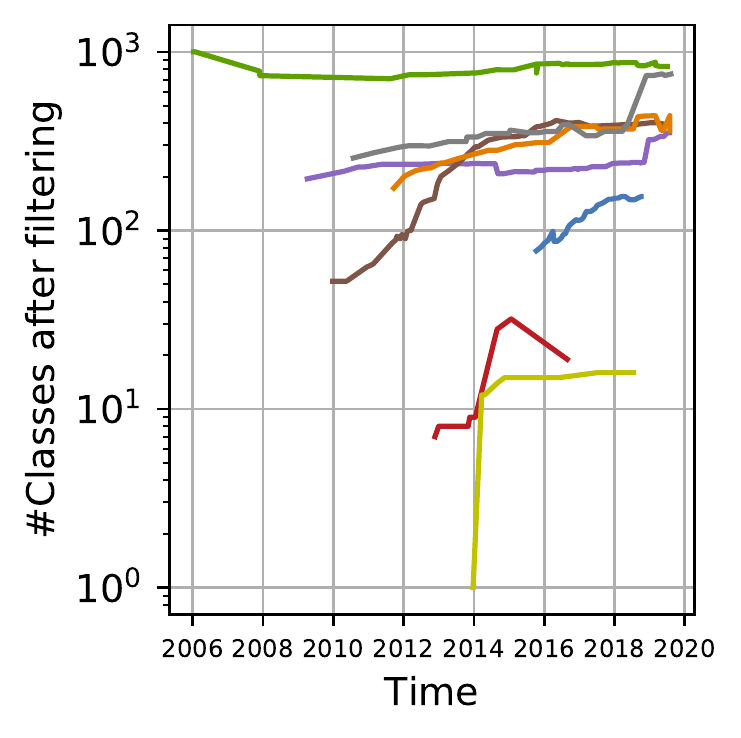}
		\vspace{-2em}
		\caption{}
        \label{fig:cleaned_class_count}
    \end{subfigure}
	~
    \begin{subfigure}[t]{0.24\textwidth}
            \centering
            \includegraphics[trim={0.25cm 0 0.3cm 0},clip,width=\textwidth]{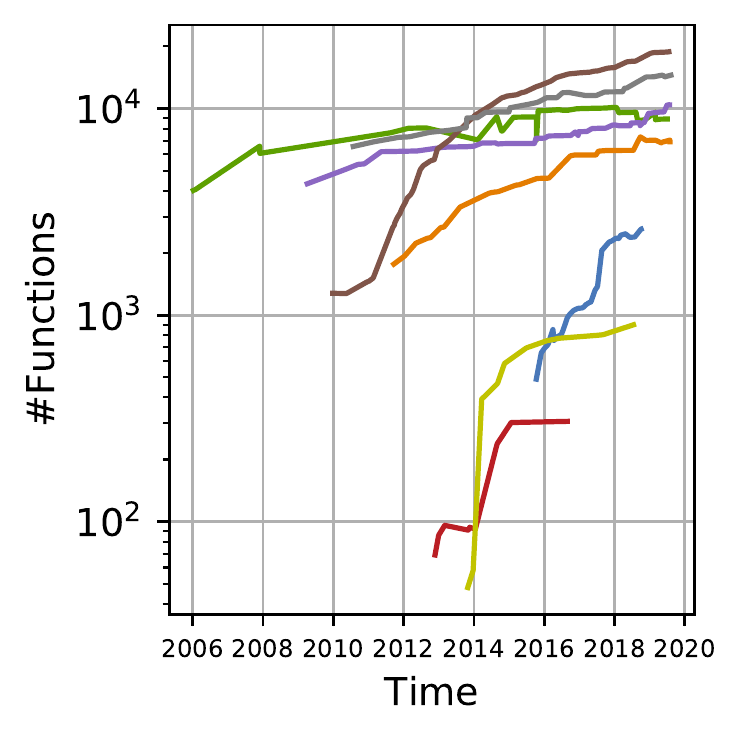}
            \vspace{-2em}
            \caption{}
            \label{fig:function_count}
    \end{subfigure}
	~
    \begin{subfigure}[t]{0.24\textwidth}
          \centering
           \includegraphics[trim={0.25cm 0 0.3cm 0},clip,width=\columnwidth]{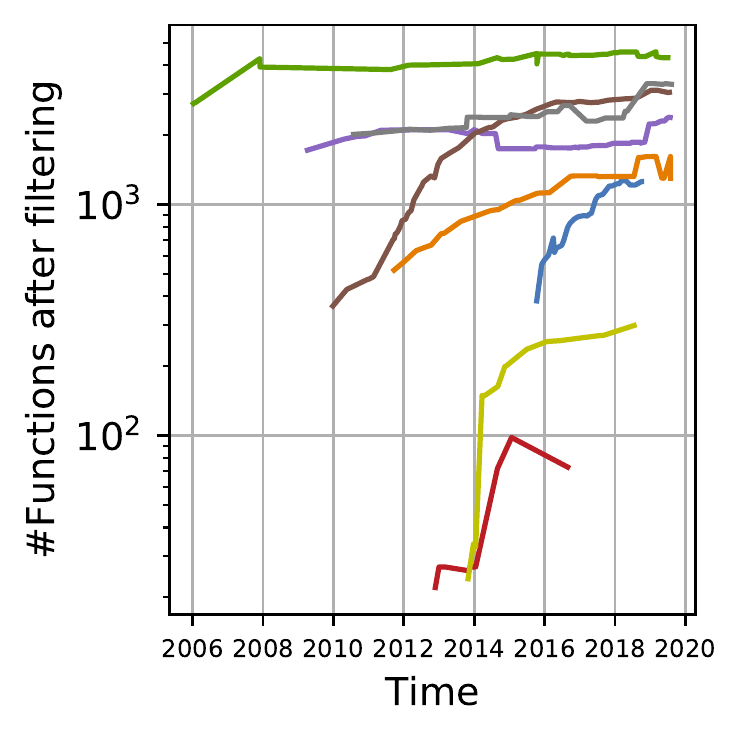}
           \vspace{-2em}
           \caption{}
        	\label{fig:cleaned_function_count}
    \end{subfigure}
    \vspace{-1em}
    \caption{Number of Python classes (a), classes after filtering (b), functions (c), and functions after filtering (d) per library over time. (The legend in (a) serves as the legend  for all (a), (b), (c), and (d).) }
    \vspace{-1em}
\end{figure*}

\textbf{Classes.} \autoref{fig:class_count} shows the distribution of Python classes per library over time. We make three main observations. First, similarly to files, we observe that every library has gone (or is going) through phases of rapid increase, this time in terms of Python classes. Second, we can see that, for most of the libraries, the number of classes is increasing steadily and at significantly different scales. For example, \matplotlib, \scipy, \numpy, \pandas, and \sklearn have a lot more classes than other libraries. Third, in contrast to files, classes tend to experience lower variance in their volume over time. Finally, note that classes in \autoref{fig:class_count} include both classes used for testing or example purposes, as well as classes that are part of the library APIs.

\textbf{Classes after filtering.} To get a better understanding of classes exposed as part of the API of libraries, we removed classes (a)~that appear in files under directories whose name contains one or more of the strings ``examples'', ``test'', and ``doc''; and (b)~whose name either begins with \_ (in Python such classes are considered hidden) or contain one or more of the strings ``examples'', ``test'', and ``doc''. Upon removal, the number of classes per library over time is shown in \autoref{fig:cleaned_class_count}. Comparing this with the unfiltered counts leads us to two interesting observations. First, the reduction on the number of classes ranges from \mysim13\% and \mysim31\% for \keras and \matplotlib, respectively, to \mysim74\% and \mysim79\% for \seaborn and \numpy, respectively. These results indicate that these libraries are getting thoroughly tested and provide many examples and documentation of their code. \wag{We believe that this is a good reason for their high usage in the data science projects}. Second, it becomes more clear that \matplotlib and \scipy have the most classes; 1,004 and 753 classes, respectively.

\textbf{Functions.} Besides classes, we also performed the same analysis on functions.~\autoref{fig:function_count} shows the distribution of functions for every library over time. We make two main observations. First, the results are similar to the ones on classes, and what we discussed as observations on the growth of classes also apply on functions. Second, however, we observe that the number of functions of every library is larger than the number of classes (typically one order of magnitude more). Finally, it is interesting to see that the relative order of libraries with respect to the number of functions and classes remains similar. Yet, we note that \pandas and \scipy have significantly more functions than \matplotlib and \numpy, which was not the case for classes.

\textbf{Functions after filtering.} Finally, similarly to classes, we performed an analysis by removing hidden functions and functions related to tests, documentation, and examples from the set of functions above. \autoref{fig:cleaned_function_count} shows the distribution of Python functions per library over time. After filtering, one of the most important observations is that again there is a big reduction, this time in number of functions. This reduction results in \matplotlib becoming the library with the most functions, while \pandas and \scipy now drop to second and third place in this ranking. These results are again indicative of the efforts of making example/test suites and documentation by the respective communities. Furthermore, the (often exponential) increase in the number of functions per library, although it has certainly decreased over time for many libraries, it is by no means indicative of APIs that have reached consensus. \wag{Hence, we speculate that systems for ML that aim to manage and optimize the functionality of these libraries need to account for potential additions or deprecations over time.}

\eat{
In this section, we look at individual packages in more details. \autoref{ss:version} discusses the release of each packages in the past decade.  
}

\eat{
\autoref{fig:class_count},~\ref{fig:function_count}, and~\ref{fig:file_count} show the number of classes, functions, and files for each package under different releases. We can see that for most of the packages, the numbers of functions and classes are increasing steadily and with significant different scales, e.g. \texttt{matplotlib}, \texttt{scipy}, \texttt{numpy}, \texttt{pandas}, and \texttt{scikit-learn} have a lot more functions and classes than the other packages. Te number of functions and classes for \texttt{keras} increased a lot in 2014. The y-axis is in log-scale, which means a steady increase in the figures indicates an exponential increase.
}

\eat{
\subsection{Library Version Analysis}
\label{ss:version}
}

\eat{
\begin{figure}[h!]
	\centering
	\includegraphics[width=\columnwidth]{list1_pypi_new.pdf}
	\caption{Release frequency of \texttt{numpy}, \texttt{matplotlib}, \texttt{pandas}, \texttt{scipy} and \texttt{seeaborn}}
\end{figure}

\begin{figure}
	\centering
	\includegraphics[width=0.5\textwidth]{list2_pypi.pdf}
	\caption{Release frequency of machine learning packages}\label{fig:version2}
\end{figure}

\begin{figure}
	\centering
	\includegraphics[width=0.5\textwidth]{list3_pypi.pdf}
	\caption{Release frequency of \texttt{nimbusml} and \texttt{mxnet}}\label{fig:version3}
\end{figure}
}

\eat{
\ifp{@Yiwen, can you verify the below?}
}

\eat{
\AF{Maybe use another term instead of clean? Maybe filtered?}
\fp{I changed it to classes/functions after filtering. I think filtered implies that we depict the filtered out ones.}
}

\eat{
\kk{The last period seems disconnected. In the one before you say that stable num of files does not mean stable functionality. Do you want to say here that non-stable number of files does not mean the opposite either? Still if you change documentation, you'd change functions too, so it would be unstable. Maybe for testing it is more true, but still. I'd keep just the first comment for stable files num does not mean stable functionality. The rest becomes more clear below.}

\fp{Documentation may not be tight to functions. It may be just be documentation for the system in general. Also increase in testing files may just mean you increased coverage from testing without changing the functionality (functions/classes). So, if there is stability in #files does not imply stability in functionality. Hence, we dive deeper in functions and classes.}

\fp{Changed the text slightly, to account for this.}
}

\eat{
\AF{We need to say why we picked these libraries? Are they the most popular? Point to apreviosu figure?}
\fp{handled}
}

\eat{
\AF{In the graphs why for some libraries we dont have data after some time? E.g nolearn after 2016? Should we mention somewhere why is that?}
\fp{Because there is no release after that time}
}

\eat{
\begin{figure}[t!]
	\centering
	\includegraphics[width=.9\columnwidth]{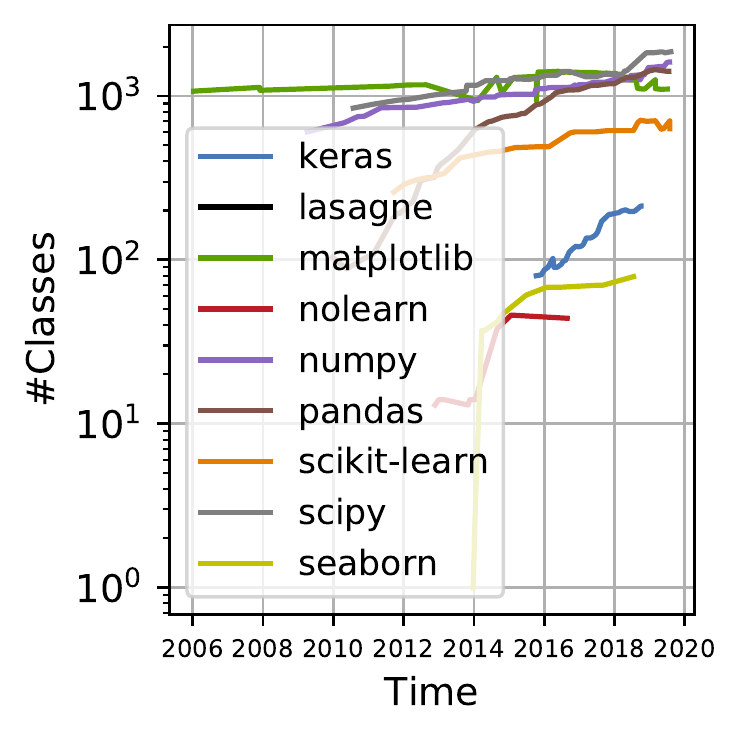}
	\vspace{-1.5em}
	\caption{\#Python classes per library over time.}
	\vspace{-6mm}
\end{figure}
}

\eat{
\begin{figure}[t!]
	\centering
	\includegraphics[width=.9\columnwidth]{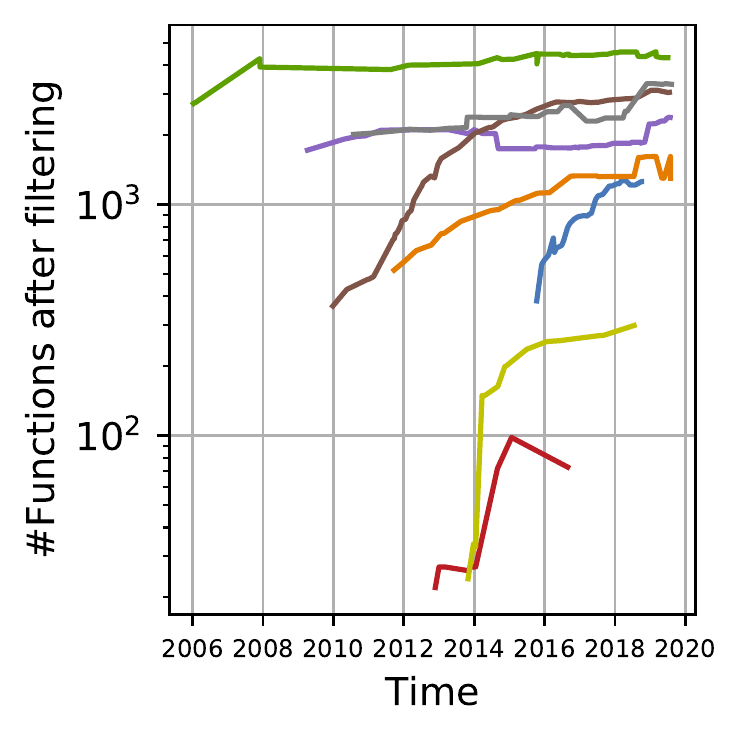}
	\vspace{-5mm}
	\caption{\#Python functions after filtering per library over time.}
	\vspace{-6mm}
\end{figure}
}

\eat{
\begin{figure}[h!]
	\centering
	\includegraphics[width=.9\columnwidth]{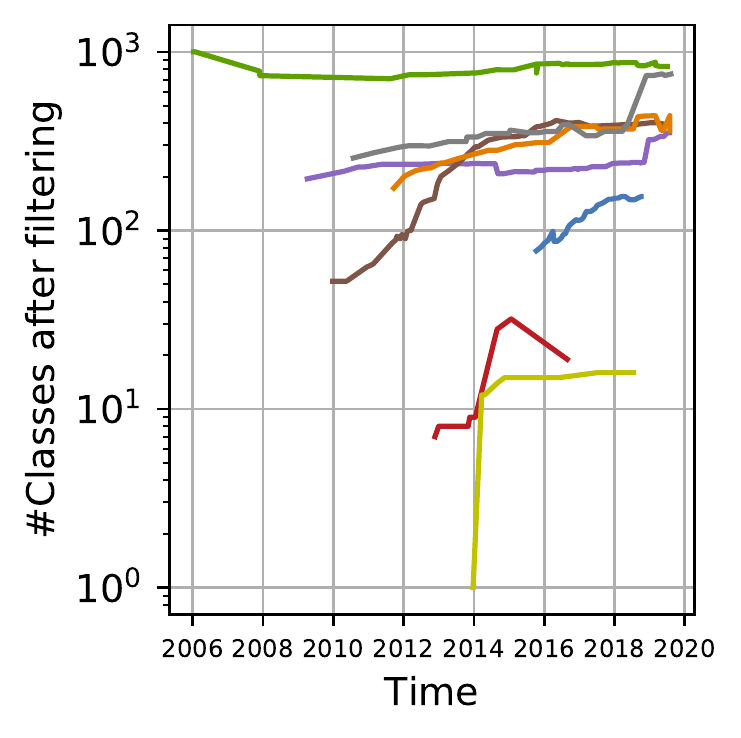}
	\vspace{-1.5em}
	\caption{\#Python classes after filtering per library over time.}
	\vspace{-1em}
\end{figure}
}
\eat{
\begin{figure}[t!]
	\centering
	\includegraphics[width=.9\columnwidth]{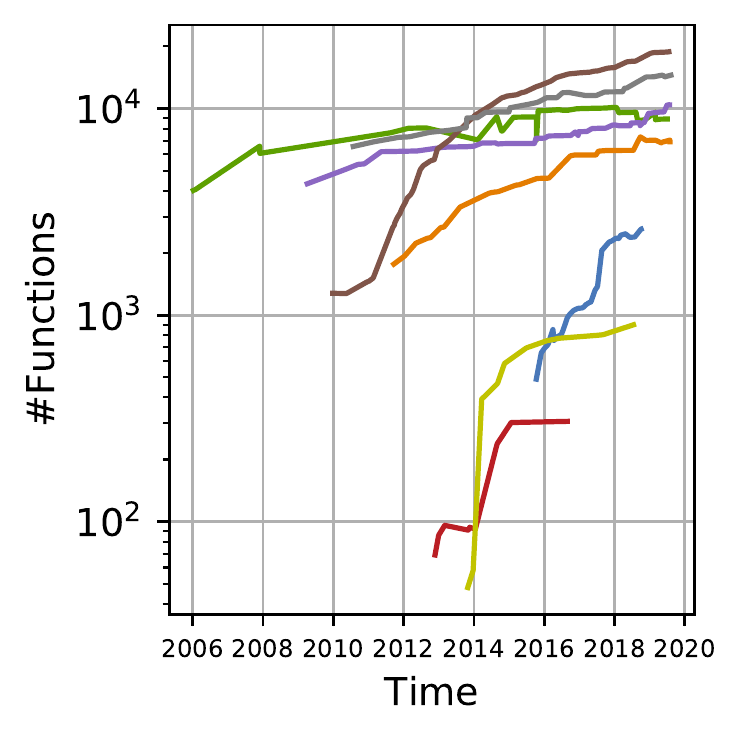}
	\vspace{-5mm}
	\caption{\#Python functions per library over time.}
	\vspace{-6mm}
\end{figure}
}

\eat{
\begin{figure}[t!]
	\centering
	\includegraphics[width=.9\columnwidth]{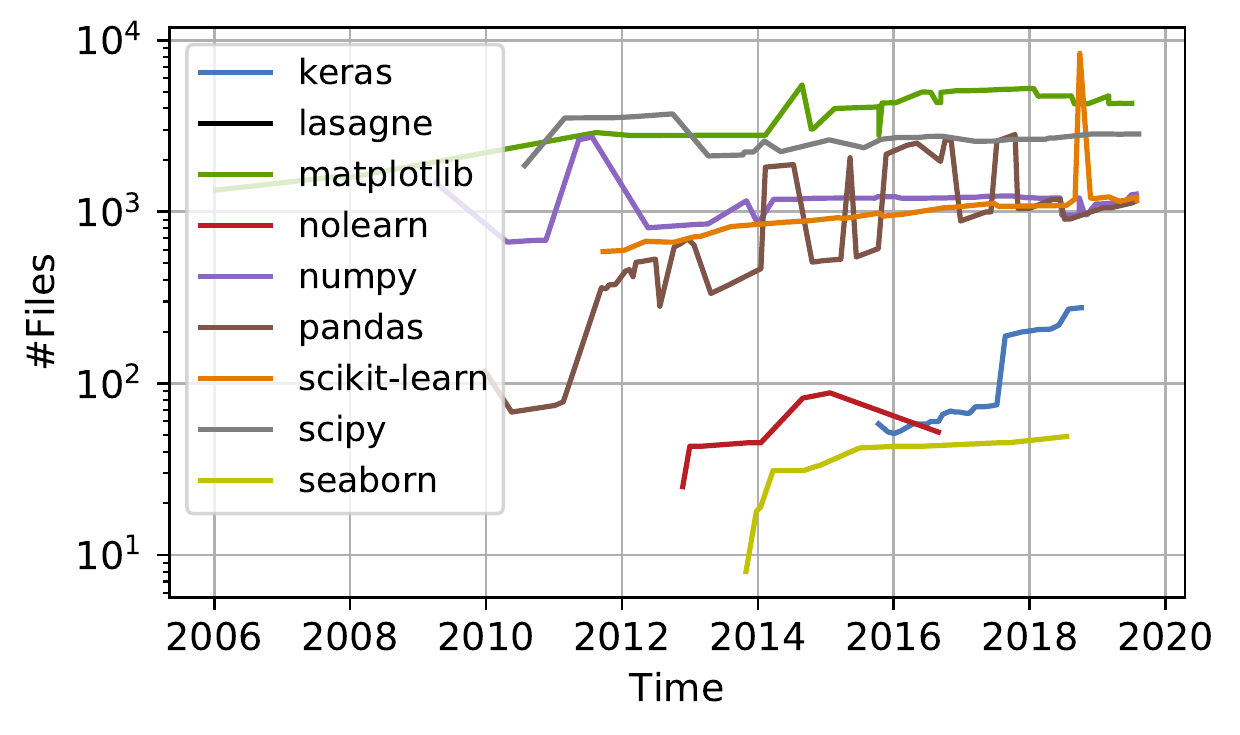}
	\vspace{-1.5em}
	\caption{\#files per library over time.}
	\vspace{-2em}
\end{figure}
}

\eat{
\begin{figure}[t!]
	\centering
	\includegraphics[width=.9\columnwidth]{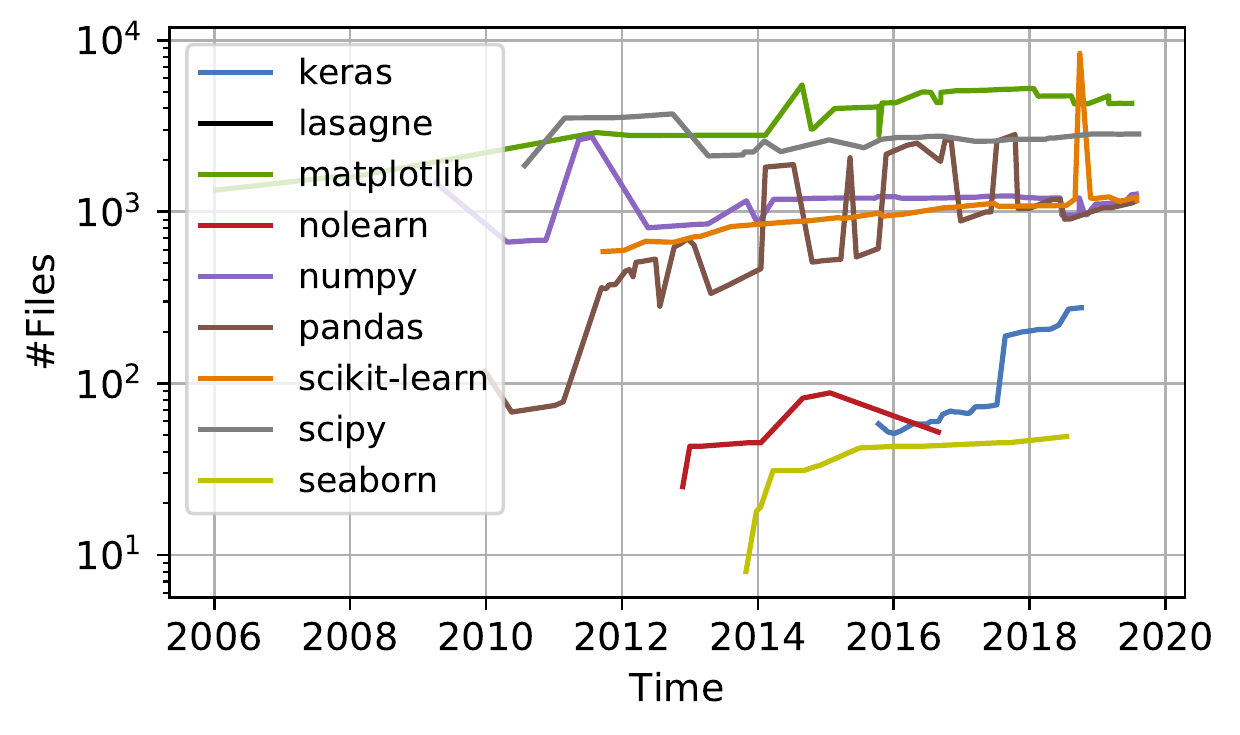}
	\vspace{-1.5em}
	\caption{\#files per library over time.}
	\vspace{-2em}
\end{figure}
}

\eat{
\begin{figure}[t!]
	\centering
	
	\vspace{-4mm}
	\caption{}

	\vspace{-4mm}
\end{figure}

\begin{figure}[t!]
	\centering
	\vspace{-4mm}
	\vspace{-1.5em}
\end{figure}
}

\section{Related Work}
\label{s:related}

Understanding data science processes is crucial as most applications are becoming ML-infused~\cite{cidrvision}. Our work sheds some light on the topic by performing an analysis of Python notebooks and \tlc pipelines. Other approaches~\cite{amershi2019software} include extensive discussions with data scientists about their software engineering practices. 
 
The work in ~\cite{pypianalysis} presents a coarse-grained analysis on the PyPI repository. Our work targets the Python language as well but with a special focus on data science applications. Thus, we look at a much larger corpus (including PyPI) and provide fine-grained analysis on data science related packages. The work in~\cite{decan} compares the package dependency graphs of CRAN, PyPI and NPM. This work targets various languages and thus does not contain a detailed analysis on Python packages. The study in~\cite{rule:2018:collection} performs an analysis of GitHub notebooks with a focus on the interaction between exploration and explanation of results. Our work incorporates the dataset used for this study (\ghold) but focuses more on the structure of data science code rather than the usage of notebooks when explaining the results of the analysis.

\section{Conclusions}
\label{s:conclusions}

Machine Learning is becoming an ubiquitous technology, leading to huge engineering and research investments that are quickly reshaping the field. As builders of ML infrastructure and DS practitioners, we felt the need for a better vantage point on this  shifting panorama. We thus amassed a large amount of DS projects (tallest pile to date to the best of our knowledge), and climbed it by means of static analysis, and statistical characterization. From this vantage point, we make several immediate observations and to inform our future work we dare to guess what's happening further in the distance. This analysis, like any other, has several limitations, but we believe is pragmatically very useful. This paper, and future releases of the underlying data and tools, is an invite to the community to join us in enjoying this view, so that we can have a common understanding on the space we all operate in.

\clearpage
\balance

\bibliography{main}
\bibliographystyle{sysml2019}

\eat{
\clearpage
\appendix

}

\end{document}